%% file: main.tex
\documentclass[10pt,twocolumn,letterpaper]{article}

\usepackage[pagenumbers]{wacv} 

\usepackage{graphicx}
\usepackage{amsmath}
\usepackage{amssymb}
\usepackage{amsfonts}
\usepackage{mathtools}
\usepackage{bbm}
\usepackage{booktabs}
\usepackage[T1]{fontenc}
\usepackage{tabularx}
\usepackage[accsupp]{axessibility}  

\DeclareMathOperator*{\argmin}{arg\,min}

\usepackage[pagebackref,breaklinks,colorlinks]{hyperref}

\usepackage[capitalize]{cleveref}
\crefname{section}{Sec.}{Secs.}
\Crefname{section}{Section}{Sections}
\Crefname{table}{Table}{Tables}
\crefname{table}{Tab.}{Tabs.}

\usepackage{xcolor,colortbl}

\definecolor{Gray}{gray}{0.85}
\definecolor{LightCyan}{rgb}{0.88,1,1}

\usepackage{algorithm}
\usepackage{algorithmic}

\def\wacvPaperID{766} 
\def\confName{WACV}
\def\confYear{2024}

\begin{document}

\title{REALM: Robust Entropy Adaptive Loss Minimization for Improved Single-Sample Test-Time Adaptation}

\author{Skyler Seto, Barry-John Theobald,  Federico Danieli, Navdeep Jaitly, Dan Busbridge\\
Apple\\
{\tt\small \{sseto, barryjohn\_theobald, f\_danieli, njaitly, dbusbridge\}@apple.com}
}

\maketitle

\begin{abstract}
   Fully-test-time adaptation (F-TTA) can mitigate performance loss due to distribution shifts between train and test data (1) without access to the training data, and (2) without knowledge of the model training procedure.  In online F-TTA, a pre-trained model is adapted using a stream of test samples by minimizing a self-supervised objective, such as entropy minimization.  However, models adapted with online using entropy minimization, are unstable especially in single sample settings, leading to degenerate solutions, and limiting the adoption of TTA inference strategies. Prior works identify noisy, or unreliable, samples as a cause of failure in online F-TTA.  One solution is to ignore these samples, which can lead to bias in the update procedure, slow  adaptation, and poor generalization.  In this work, we present a general framework for improving robustness of F-TTA to these noisy samples, inspired by self-paced learning and robust loss functions.  Our proposed approach, Robust Entropy Adaptive Loss Minimization (REALM), achieves better adaptation accuracy than previous approaches throughout the adaptation process on corruptions of CIFAR-10 and ImageNet-1K, demonstrating its effectiveness.  
\end{abstract}

\input{intro}

\input{background}

\input{rw}

\input{method}

\input{experiments}

\input{conclusion}

\section*{Acknowledgements}
We are grateful to Zakaria Aldeneh, Masha Fedzechkina, and Russ Webb for their helpful discussions, comments, and thoughtful feedback in reviewing this work.

\newpage
{\small
\bibliographystyle{ieee_fullname}
\bibliography{egbib}
}


\clearpage
\appendix
\input{appendix2}

\end{document}

%% file: intro.tex
\section{Introduction}
\label{sec:intro}

Deep Neural Networks (DNNs) can achieve excellent performance when evaluated on data from the same distribution used to train the model.  However, after model deployment, natural variations, sensor degradation, or changes in the environment can cause test samples to appear different from the samples used to train the model. Such distribution shifts, may significantly worsen performance \cite{hendrycksbenchmarking}.

Test-time adaptation (TTA) is a strategy used to counter distribution shifts during online evaluation/model deployment. TTA updates the model parameters within the inference procedure through self-supervision.   In the \emph{Fully-TTA} (F-TTA) setting, the goal is to adapt an arbitrary pre-trained model on test data without access to the original training data (also called source-free), without supervision, and without access to changing the way the model was trained \cite{wangtent}.  F-TTA is an important solution for tackling source-free distribution shifts when: (1) models are deployed and source data are not available, e.g., for privacy concerns, (2) it may be prohibitively expensive to re-train the models, and (3) data from the unseen distribution may not be available immediately, and it is infeasible to wait for enough data to annotate and train with supervision.

\begin{figure*}
  \centering
  \begin{subfigure}{0.24\linewidth}
    \includegraphics[width=\columnwidth]{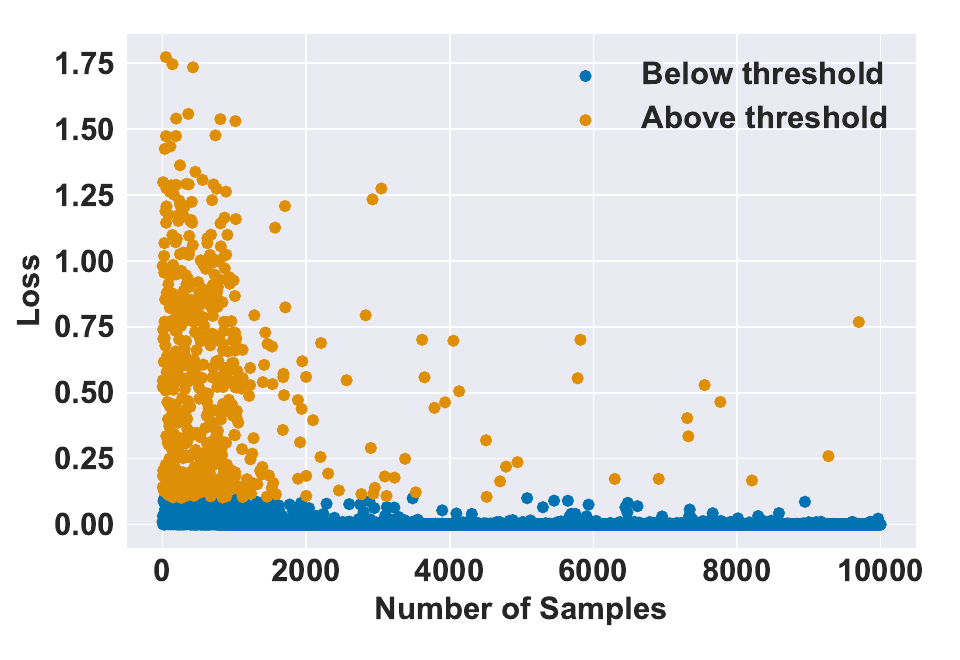}
    \caption{TENT  Loss}
  \end{subfigure}
  \hfill
  \begin{subfigure}{0.25\linewidth}
    \includegraphics[width=\columnwidth]{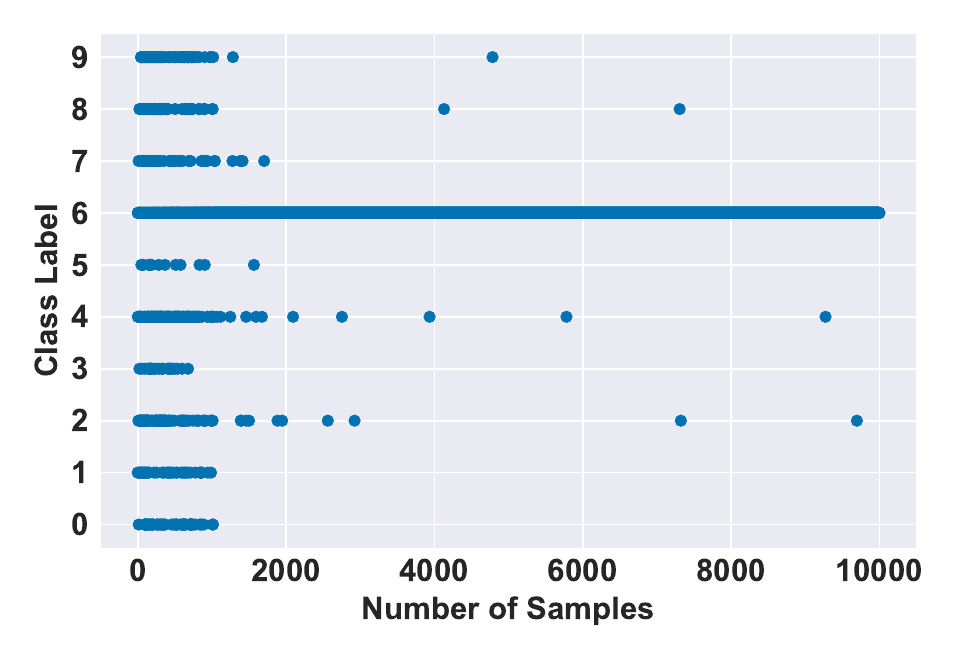}
    \caption{TENT  Predicted Class}
  \end{subfigure}
  \begin{subfigure}{0.24\linewidth}
    \includegraphics[width=\columnwidth]{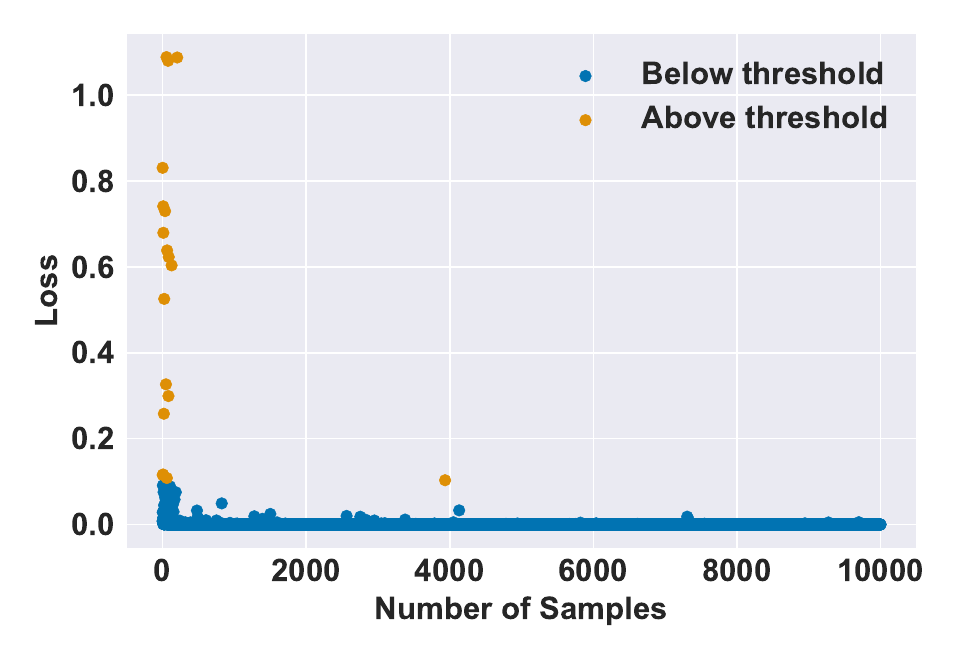}
    \caption{MEMO  Loss}
  \end{subfigure}
  \hfill
  \begin{subfigure}{0.25\linewidth}
    \includegraphics[width=\columnwidth]{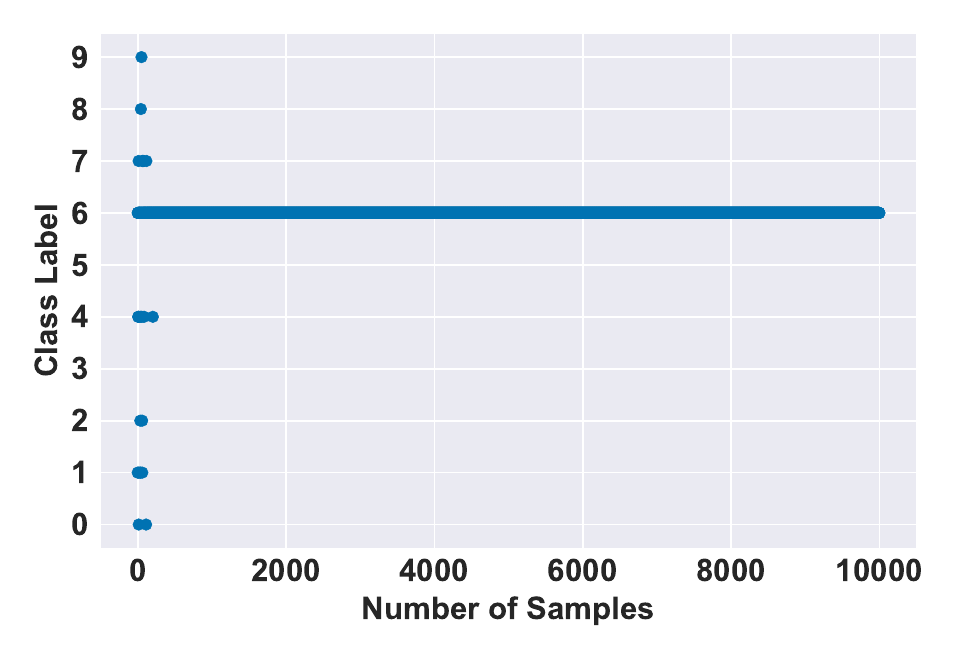}
    \caption{MEMO  Predicted Class}
  \end{subfigure}
  \caption{Loss and predicted class labels for both TENT and MEMO on the CIFAR-10 test set under gaussian noise corruption at severity 5.  Both adaptation strategies are applied to a ResNet-26 with group norm layers using standard hyperparameters from \cite{wangtent} and \cite{zhang2022memo}, but in the online adaptation setting and a batch size of 1. (a), (c): Certain samples have high loss at th start of training.  (b), (d): Model predicted classes quickly become the same across all inputs.}
  \label{fig:overfit}
\end{figure*}

A common paradigm in F-TTA is training models to minimize the entropy of the predictions \cite{wangtent}, which is typically done in two ways: (1) episodic: where models are updated and reset after each batch of samples, and (2) online: where weights are not reset after each test sample allowing for update accumulation \cite{lee2022surgical}. While several papers have examined F-TTA \cite{fleuret2021test, gao2023back, khurana2021sita, iwasawa2021test, nado2020evaluating, niu2022efficient, niu2023towards, schneider2020improving, wangtent, zhang2022memo} for online adaptation, many methods are sensitive to known issues: (1) batch normalization running statistics updating on small number of samples from the new distribution\cite{niu2023towards}, (2) noisy samples with high entropy leading to unstable model updates\cite{niu2022efficient, niu2023towards}, and (3) hyper-parameter sensitivity causing models to shift too much from the source model\cite{zhao2023pitfalls}.  Furthermore, online F-TTA has been shown to yield worse performance as the amount of adaptation data increases, and to be prone to over-fitting  leading to degenerate solutions \cite{lee2022surgical, niu2023towards}.  Additionally, little is known about online F-TTA performance in extreme scenarios with a limited number of adaptation steps, and limited amount of data per adaptation step (i.e., batch size of one for real-world inference deployment\footnote{A natural example might be an individual taking picture(s) on their phone on a rainy day. A phone app providing object recognition capabilities would be expected to classify on the new distribution given only a few samples, which are processed individually.}).

In this work, we examine the reliability of current approaches for online F-TTA with limited data \emph{and} limited adaptation step settings.
It is important to have \emph{both} of these conditions, as restricting to \emph{only} a limited number of adaptions does not constrain optimization, 
providing the stochastic differential equation (SDE) approximation of the optimization is valid\footnote{It is likely that the SDE approximation holds in the TTA setting as TTA is performed in the low-LR, small batch size regime, which corresponds to low discretization \cite{DBLP:conf/nips/LiMA21}.} \cite{DBLP:conf/nips/LiMA21}.
Equivalently, restricting to a small number of samples per adaption step does not constrain the optimization,
whereas restricting to a small number of total samples does.

Our main contributions are (1) We highlight that with batch size of one\footnote{In this work, we refer to this setting as single sample TTA.  While we adapt using the entire evaluation set, seeing samples individually still presents a challenge as the model update over a single sample can be very noisy and unreliable.}, online F-TTA can fail to adapt to the target distribution and instead quickly moves toward a degenerate solution (predicting the same label for all samples) due to high loss sample. 
(2) We study procedures which perform stable online F-TTA by skipping samples according to a reliability criteria \cite{niu2022efficient, niu2023towards}, which identify that unreliable noisy samples are a cause of unstable adaptation, and characterize their objective as part of a broader objective relating to Self-Paced Learning (SPL) \cite{fan2017self, kumar2010self}. 
(3) We propose a general variant of the SPL framework for online TTA called Robust Entropy Adaptive Loss Minimization (REALM) that updates using samples scaled by a robust loss function. 

Empirically, REALM yields better performance at early stages of adaptation (few adaptation steps/limited samples), and leads to better performance over the full test set compared to related entropy minimization methods.

%% file: background.tex
\section{Reliable Test Time Adaptation}
\label{sec:tta}
In this section, we give a brief overview of test time adaptation, show that strategies based on entropy minimization fail to adapt to the distribution shift due to noisy samples, and  define prior work that skip samples for stable TTA.

\subsection{Overview of Test Time Adaptation}
Let $f(\cdot \, ; \theta)$ be a model trained on a training set $\mathcal{D}_{\textrm{train}} = \{(x_i, y_i) \sim P_{\textrm{train}}\}_{i=1}^N$.  The goal of TTA is to  improve the performance of $f(\cdot \, ; \theta)$ on the evaluation of a test distribution $P_{\textrm{test}}$, where $P_{\textrm{train}} \neq P_{\textrm{test}}$, without access to how $f(\cdot\, ; \theta)$ was trained, and without access to $\mathcal{D}_{\textrm{train}}$.   In practice when optimizing over the test set, we have access to the batch of samples $x$ without corresponding labels $y$, that is $\mathcal{D}_{\textrm{test}} = \{(x_i)_{i=1}^M \sim Q\}$. In TTA, the model parameters $\theta$ are adapted by batchwise minimizing over the test data:
\begin{equation}
    \theta^* =  \min_{\theta}\mathcal{L}_{\textrm{SSL}}(\theta; \mathcal{D}_{\textrm{test}}),
\end{equation}
where $\mathcal{L}_{\textrm{SSL}}$ is some self-supervised objective, such as minimizing entropy \cite{wangtent}, or marginal entropy \cite{zhang2022memo}.  The goal of TTA is to adapt the model by optimizing the above objective in an online inference setting, where batches of data are streamed to the model, and predictions are made on-the-fly. This setting is challenging, especially when the amount of data at each step is limited.  The model is expected to perform well immediately in the adaptation phase, yet adaption might involve only a small number of updates to keep inference time efficient, and data might not be stored (e.g., for sensitive data with privacy considerations, or limited storage devices).  \emph{Note that in the online setting no termination is necessary, however in episodic settings, the model can be reset after new or no data appears.}  

\begin{figure*}
    \centering
    \includegraphics[width=\textwidth]{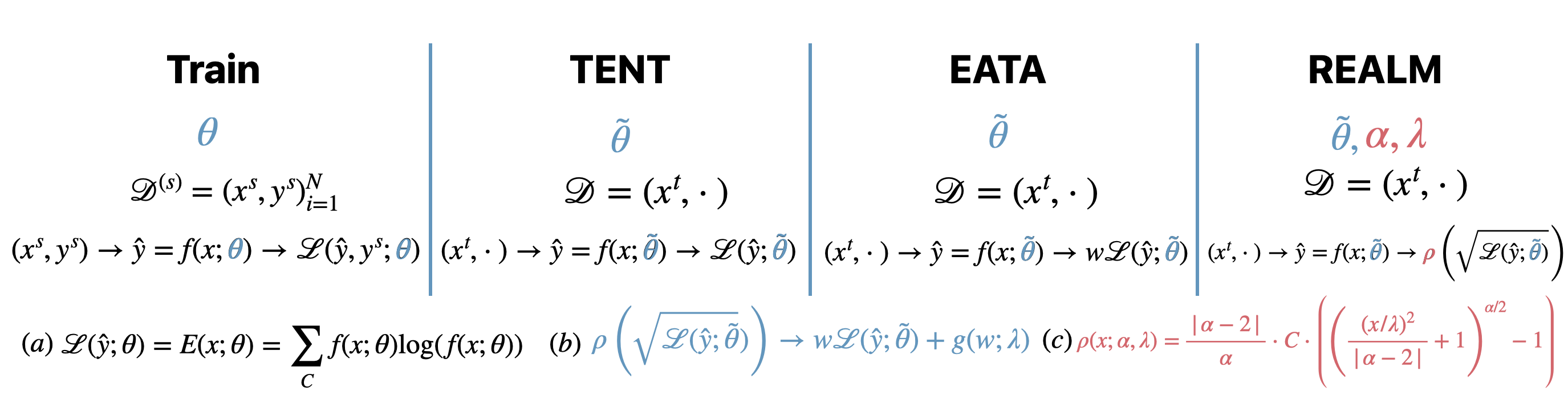}
    \caption{An overview of REALM and similar methods. For TTA, the model $f(\cdot \, ; \theta)$ training procedure stays fixed, while inference changes. (a) TENT proposes batch-wise entropy minimization of all samples, while EATA minimizes the entropy of only samples with low entropy according to the weight $w = S(x)$. We formalize optimization procedure of EATA as (b) an instance of Self-Paced Learning. Our algorithm REALM, minimizes (c) a robust loss function $\rho$ of the entropy to stabilize online adaptation against outliers.  We show that our proposed robust loss function generalizes (b).}
    \label{fig:overview_realm}
\end{figure*}

\subsection{Limitations of Test Time Adaptation}

In online adaptation scenarios with limited batch sizes, we observe that a model may quickly overfit to the self-supervised objective $\mathcal{L}_{\textrm{SSL}}$, also known as entropy collapse, resulting in performance degradation as the model maximizes its confidence by predicting the same class label. Figure~\ref{fig:overfit} illustrates this effect with two baseline TTA strategies: TENT \cite{wangtent} and MEMO \cite{zhang2022memo}, which minimizes entropy irrespective of the sample. When using each test sample to adapt the model, some samples seen early in adaptation have high loss (\cref{fig:overfit}), and when these samples are used to adapt the model, the model learns to always predict the same class label.  Concurrent work, SAR, identified a similar phenomena with batched TENT through comparison of the entropy and gradient norms on ImageNet with gaussian noise corruption at severity 5 \cite{niu2022efficient}.  

To mitigate entropy collapse, recent works propose re-weighting the entropy objective with an indicator function on the sample entropy that results in skipping updates for high entropy (i.e., \emph{unreliable}) samples\cite{niu2022efficient, niu2023towards}. The EATA approach \cite{niu2022efficient} introduces the objective:
\begin{equation}
    \min_{\theta} S(x) \,\mathcal{L}(\theta; x).
    \label{eq:eata}
\end{equation}
The particular weight $S(x)$ is written in two parts:
\begin{equation}
    S(x) = S_{\textrm{ent}}(x)\cdot S_{\textrm{div}}(x),
\end{equation}
\begin{equation}
     S_{\textrm{ent}}(x) = \frac{1}{\exp\left[\mathcal{L}(\theta) - \lambda\right]}\cdot \mathbbm{1}\left\{\mathcal{L} < \lambda\right\},
     \label{eq:eata_weight}
\end{equation}
for some pre-defined threshold on the entropy, $\lambda$. Such a procedure has also been used for building robust optimizers \cite{shah2020choosing}. $S_{\textrm{div}}$ is defined as:
\begin{equation}S_{\textrm{div}} = \mathbbm{1}\left\{\cos\left(f(x; \theta), m_{t-1}\right)\ < d \right\}, \end{equation}
where $m_{t-1}$ is an Exponential Moving Average (EMA) of the predictions from prior updates\footnote{Note that $S_{\textrm{div}}$ is operationalized as a {\tt torch.where()} and a stop gradient is applied such that no update is done even though $S$ is a function of $\theta$.  Thus, we can simply think of $S(x)$ as a function of the sample only.}.

However, as EATA is an online procedure, when samples are seen only once, some samples may never be used to update the model, especially with a batch size of one, which can lead to class-imbalance and potential bias in the update procedure. Further, this process results in few updates at the start of testing, which results in slow adaptation to the shifted distribution, and a dependency on the ordering of samples during adaptation.  Concurrent to this work, SAR \cite{niu2023towards} extends EATA by also modifying the optimization procedure with Sharpness Aware Minimization (SAM) \cite{foretsharpness}, which encourages the model to move towards flat entropy. An overview of F-TTA approaches such as Tent and EATA\footnote{We refer to the online F-TTA setting in this work as just TTA as  all comparisons made all primarily operate under the online Fully TTA setting. We specify when comparing other methods and in the related work when such methods operate under a different setting.} are shown in \cref{fig:overview_realm}.  The primary differenec between our approach REALM and prior works is that we modify the entropy minimization objective using a robust loss function $\rho$, a generalization of the weighting approach in EATA, and Tent which induces entropy collapse.

%% file: rw.tex
\section{Related Work}
\label{sec:rw}

The goal of this work is to adapt a pre-trained model trained on (source) data using only unlabeled data from a new (target) distribution at test time.  In the online setting, the data are assumed to arrive in a streaming fashion, and can be used only once to update the model.  This is in contrast to similar fields, such as \emph{domain adaptation} \cite{kouw2019review}, which trains a model with data from the training distribution (source) and unlabeled data from the test distribution (target), and \emph{domain generalization} \cite{zhou2022domain}, which trains a model on multiple domains to generalize to new domains at test time.   Although there are source-free methods within the \emph{domain adaptation} literature \cite{kundu2020universal, li2020model, liang2020we}, methods that modify the training procedure for better TTA such as Test-time training \cite{d2020one, liu2021ttt++, sun2020test}, and methods that carry an additional model, such as a diffusion model trained on source data \cite{gao2023back}, they assume access to a large amount of data from the source or target distribution for potentially multiple epochs of training, and are outside the scope of this work.  For our work, we focus on online TTA methods that minimize an entropy-based objective. 

One of the earliest works in TTA investigates adaptation of a speaker-independent acoustic model to new speakers at test-time \cite{wegmann1999dragon}.  Numerous works have since proposed different approaches for TTA in vision \cite{li2016revisiting, liu2021ttt++, schneider2020improving, sun2020test, wangtent, zhang2022memo}, and speech \cite{kim2023sgem,lin2022listen}.  Recent TTA strategies primarily focus on small updates to the model, typically only adapting the normalization layers of the network (i.e. batch normalization \cite{ioffe2015batch}, group normalization \cite{wu2018group}, and layer normalization \cite{ba2016layer}) as these layers have a small number of parameters relative to the full network, and are impacted heavily by covariate shifts \cite{benz2021revisiting, schneider2020improving}.  AdaBN \cite{benz2021revisiting, li2016revisiting} suggests updating the statistics in BN layers according to the new distribution, and \cite{schneider2020improving} suggests a rolling update of the normalziation layer statistics.  Other approaches also use variants of the batch (such as augmentations) to perform normalization statistic updates \cite{khurana2021sita}. Other methods update the parameters (momentum and scale) of the normalziation layers including \cite{niu2022efficient, niu2023towards, wangtent}.  Still, other methods selectively update parts of the full network \cite{lee2022surgical, niu2023towards} or the entire network \cite{zhang2022memo}, however apriori it is hard to know what parts of the network should be updated for an unknown distribution shift.  

Many normalization layer update methods \cite{benz2021revisiting,li2016revisiting,wangtent,niu2022efficient} rely on a large batch of samples (more than $64$) to perform adaptation, and are thus unsuitable for online TTA, especially in the single instance, or few sample regime. A recent investigation into online TTA, TENT \cite{wangtent}, minimizes the entropy of a given batch of data and continues on new batches of data.  TTT also maintains an online version, but still presupposes training with a different objective \cite{sun2020test}.  

Following on, numerous approaches have proposed unsupervised entropy-based optimizations including \cite{jing2022variational} for bayesian domain adaptation,  MEMO and TTA-PR \cite{fleuret2021test, zhang2022memo}, which optimize average or marginal entropy over multiple augmentations.  Recent works additionally investigate the instability of online TTA with entropy minimization demonstrating catastrophic drop in performance in instance-based online TTA, and poor performance from adapting on unreliable samples.    In particular, EATA \cite{niu2022efficient} suggests skipping updates on samples that have high entropy, and SAR \cite{niu2023towards} concurrent to our work notes that updating on these samples leads to overfitting the unsupervised objective, resulting in a model which always predicts the same class.  The SAR procedure uses the same weight objective as in EATA, but also uses the SAM optimizer \cite{foretsharpness}.  Other works demonstrate that limiting the parts of the model that are updated can lead to more stable online TTA \cite{lee2022surgical}.  In this work, we critically examine these approaches for improving stability in online TTA with entropy minimization, highlighting their shortcomings and proposing an improved, general framework encompassing these approaches.  

Beyond entropy minimization, pseudo-label approaches perform online TTA \cite{boudiaf2022parameter, kingetsu2022multi}, and many other attempts at attaining stability from continual learning \cite{de2021continual} including anti-forgetting regularization, and consistency regularization constrain the model parameters to be close to the source model, thus improving stability \cite{kirkpatrick2017overcoming, niu2022efficient, wang2022continual}.  However, these approaches have a different goal aimed at reducing forgetting source distribution, and not in directly stabilizing and improving adaptation to a target domain via entropy minimization objectives. Many approaches concurrent to this work also rely on memory banks \cite{yang2023auto}, storing original model weights \cite{wang2022continual}, or copies of the model \cite{song2023ecotta}. For a comprehensive survey on these approaches, and other TTA approaches see \cite{liang2023comprehensive}.

%% file: method.tex
\section{Robust Entropy Adaptive Loss Minimization (REALM)}
\label{sec:realm}

\subsection{Connecting EATA to Self-Paced Learning and Robust Loss Functions}

Consider the empirical risk minimization objective:
\begin{equation}
    \theta^* = \argmin_{\theta}\mathbbm{E}\left[\mathcal{L}_S(\theta; x)\right] = \arg\min_{\theta}
    \sum_{i=1}^N \mathcal{L}_S(\theta; x_i).
\end{equation}
Given the weight $S_{\textrm{ent}}$ from \eqref{eq:eata_weight},  we rewrite the reweighted objective \eqref{eq:eata}  through a connection to the SPL literature \cite{kumar2010self,fan2017self}.  In SPL, the aim is to solve the joint optimization problem in $w$ and $\theta$:
\begin{equation}
\begin{split}
    w^*, \theta^* 
    &= \argmin_{w, \theta}\mathbbm{E}\left[w(x)\mathcal{L}(\theta; x) + g(w; \lambda)\right],\\
    &= \argmin_{w, \theta}
    \sum_{i=1}^N [w(x_i)\mathcal{L}(\theta; x_i) + g(w(x_i); \lambda)],
\end{split}
\label{eqn:regularized-min}
\end{equation}
where $w(x_i) \in [0, 1]$ is a weight for the loss controlling the importance of the sample for learning, and $g(\cdot)$ is a regularizer on $w$ controlling the pace of learning. \\

A typical procedure for solving \cref{eqn:regularized-min} is  an alternating iterative procedure, where one first solves for the optimal weights $w^*$ holding $\theta$ fixed, then the model parameters $\theta^*$ while holding $w$ fixed.  We write EATA \cite{niu2022efficient} as a SPL objective by defining $g(w; \lambda) = -\lambda\|w\|_1$, which yields the min-min objective: 
\begin{equation}
    w^*, \theta^* = \arg\min_{w, \theta}\mathbbm{E}\left[w(x)\mathcal{L}(\theta; x) - \lambda\|w(x)\|_1\right].\\
\label{eqn:eata_spl}
\end{equation}
For this choice of $g$, its closed form solution for $w$ is $S_{\textrm{ent}}(x)$, and optimization of $\theta$ is same procedure as \eqref{eq:eata} up to some constants in $\theta$, which do not influence optimization  \cite{kumar2010self}.

We further note that for certain classes of regularizer, the SPL optimization can be written in terms of a robust loss function $\rho(x)$.  That is, optimization of the form $\min_{\theta}\sum_{i=1}^N\rho(\mathcal{L}(\theta; x_i))$  as is done in SPL with implicit\footnote{Implicit means that one need not have a closed-form expression for $g$.  Nonetheless, our choice of $\rho$ does have a closed form $g$.} regularization \cite{fan2017self}. 

For hard-thresholding, the corresponding robust loss function is the Talwar function \cite{hinich1975simple}, and for EATA, the corresponding ``robust'' loss function is
\begin{equation}
\rho_{\textrm{EATA}}(x; \lambda) = \begin{cases} 
      x & x \leq \lambda \\
       \lambda & \textrm{otherwise} ,
   \end{cases}
   \label{eqn:rhoETA}
\end{equation}
where $\lambda$ is the loss threshold and is $x$-independent.  In summary, we have shown that EATA is optimization of a weighted Talwar loss, or an SPL objective with an L1 norm on the weights.

\subsection{Robust Entropy Adaptive Loss Minimization}
One may intuitively think the corresponding loss function and weighting function need to correspond to  piecewise penalization according to the relationship between the loss threshold $\lambda$, similar to a robust loss like the Huber loss \cite{huber1992robust}. However, there are many suitable ``robust'' loss functions that are not piecewise, for example the Welsch loss function \cite{dennis1978techniques}, which yields a similar tapering on the loss, and the Cauchy loss function \cite{black1996robust}.

In this work, we suggest adaptation with a general robust loss function which interpolates many robust loss functions used in literature. To our knowledge this is the first instance of such a function applied on the entropy objective, and for TTA. The form for the adaptive loss function is written as:
\begin{equation}
    \rho(x; \alpha, \lambda) = \frac{|\alpha-2|}{\alpha}\cdot  C \cdot \left[\left(\frac{(x/\lambda)}{|\alpha-2|}+1\right)^{\alpha/2}-1\right], 
    \label{eqn:rho}
\end{equation}
for $\alpha \in (0, 2]$, and has been adapted\footnote{The original definition of the robust loss function appearing in \cite{barron2019general} results in a squared loss term as the original general robust loss is intended for squared-error loss functions.  Starting from that definition leads to a squared entropy objective. More details are in \cref{sec:realm_details}.} from \cite{barron2019general} for entropy minimization. Optimization of \cref{eqn:rho} also has the benefit of parameterizing both the shape of the loss (in terms of $\alpha$ and the threshold $\lambda$ in terms of the scale of the loss).  Thus, we suggest optimization of the general robust function of the entropy for better TTA.  This yields a far simpler procedure for stable TTA while still being robust to outliers.  Our optimization procedure, which we call \textbf{R}obust \textbf{E}ntropy \textbf{A}daptive \textbf{L}oss \textbf{M}inimization (REALM), optimizes the robust function of the entropy while simultaneously learning $\alpha$ (the shape of the loss), and threshold $\lambda$ (the scale of the loss):
 \begin{equation}
    \theta^*, \alpha^*, \lambda^* = \min_{\theta, \alpha, \lambda} S_{\textrm{div}}(x) \rho\left(\mathcal{L}(\theta; x); \alpha, \lambda\right). 
    \label{eqn:robust-min}
 \end{equation}
Further note that, under the constraint that $\alpha \in (0, 2]$, the robust minimization problem \cref{eqn:robust-min} satisfies the SPL framework, and can be recast as the regularized problem 
\begin{equation}
	\min_{\theta} \rho\left(\mathcal{L}(\theta);\alpha,\lambda\right) = \min_{\theta,w}\left[w\frac{\mathcal{L}(\theta)}{\lambda} + g(w;\alpha)\right].
	\label{eqn:regularized-robust-equivalence}
\end{equation}
The regularizer $g(w;\alpha)$ can be defined explicitly, as follows:
\begin{equation}
    g(w;\alpha) = \frac{|\alpha-2|}{\alpha}\left[w^{\frac{\alpha}{\alpha-2}}\left(1-\frac{\alpha}{2}\right) + \frac{\alpha}{2} w - 1\right].
 \end{equation}
 This formula can be obtained by considering the derivatives of \cref{eqn:regularized-robust-equivalence} with respect to $\theta$ and $w$, and by following similar considerations as in \cite{unification1996}; we refer to  \cref{sec:realm_details} for the complete derivation.

 \begin{figure}
    \centering
    \includegraphics[width=1.6in]{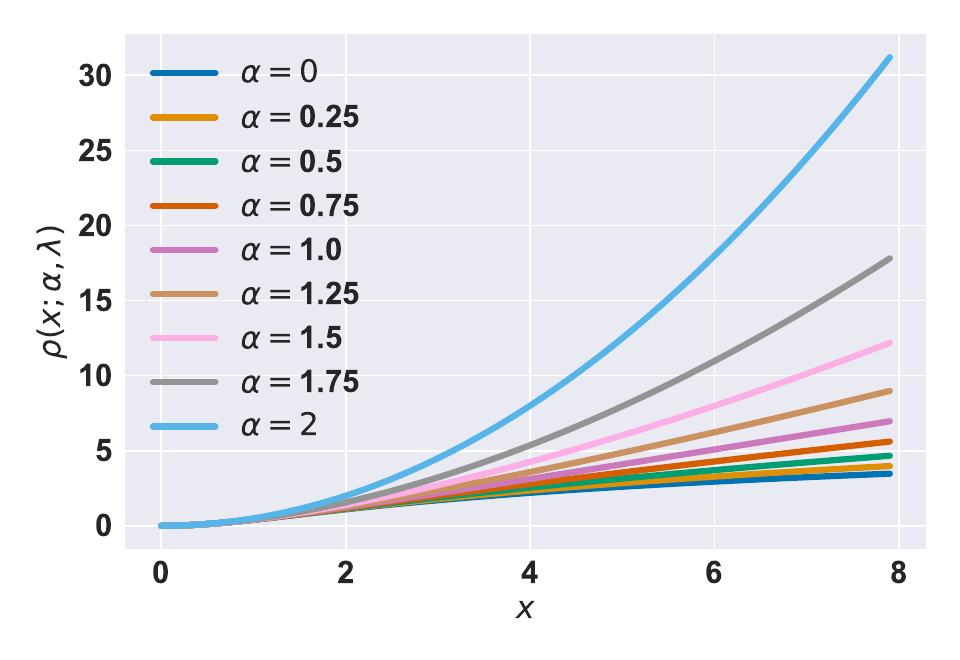}
    \includegraphics[width=1.6in]{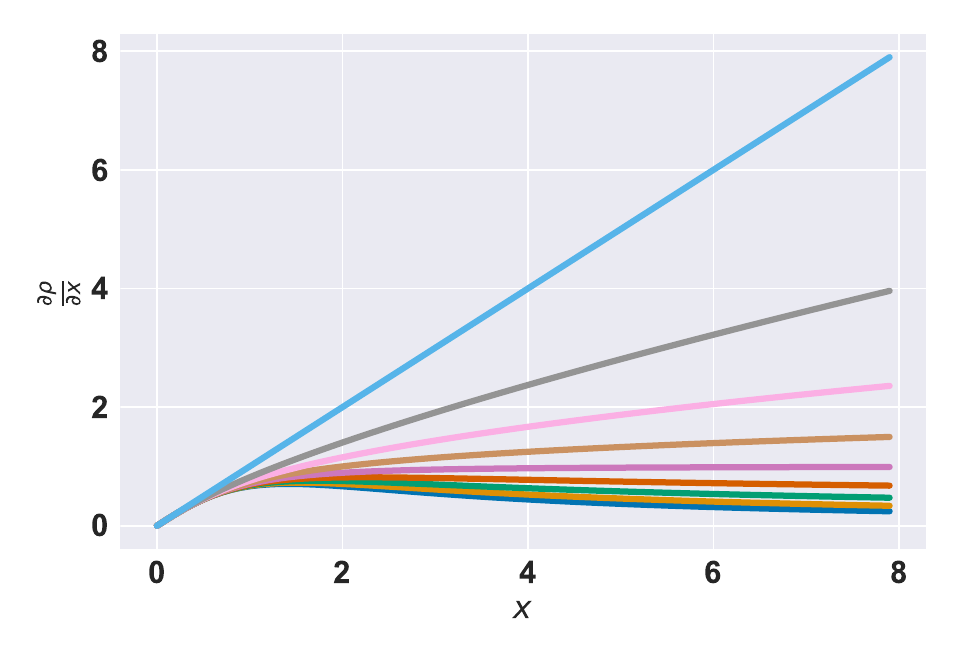}
    \caption{Robust loss (left) and its derivative (right) for varying $\alpha$ and fixed scale $\lambda=1$.}
    \label{fig:robust_loss_overview}
\end{figure}
The equivalence in \cref{eqn:regularized-robust-equivalence} implies that REALM performs TTA using a self-paced learning objective, and the theoretical motivation underpinning REALM  is the same as that for EATA, but the regularizer chosen is less strict yielding more gradient updates during optimization over EATA.

To highlight the advantage of our proposed approach, we first note that $\rho_{\textrm{EATA}}$ and $\mathcal{L}$ are both extremes on the distribution of possible scaling functions $\rho(\cdot)$.  In particular, using the standard loss results in no penalization, whereas $\rho_{\textrm{EATA}}$ is the most strict penalization resulting in no gradient update when the loss is high.  However, visualizing $\rho(x; \alpha, \lambda)$ in \cref{fig:robust_loss_overview} for the squared loss, and a range of $\alpha$ reveals many functions that still offer penalization of outliers without yielding zero gradient update on such outliers.  $\alpha \in [0, 0.5]$  yields solutions that have a small gradient update for outlier samples, while behaving similar to the initial loss for inliers.

\begin{figure*}
  \centering
  \begin{subfigure}{0.24\linewidth}
    \includegraphics[width=\columnwidth]{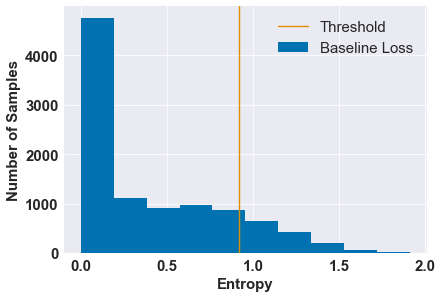}
    \caption{C10 Loss}
  \end{subfigure}
  \begin{subfigure}{0.24\linewidth}
    \includegraphics[width=\columnwidth]{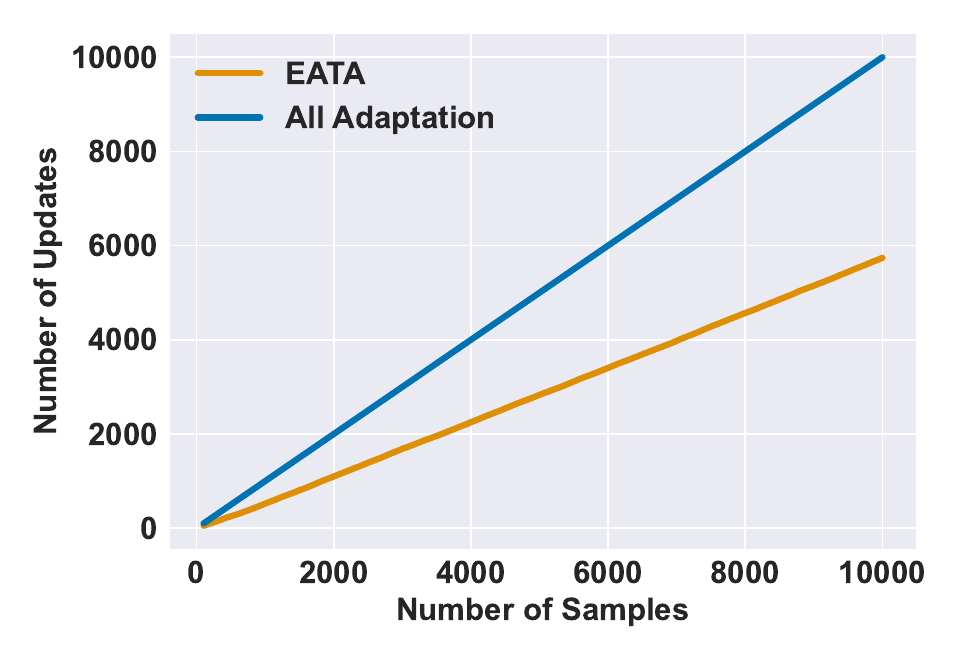}
    \caption{C10 EATA Updates}
  \end{subfigure}
  \begin{subfigure}{0.24\linewidth}
  \includegraphics[width=\columnwidth]{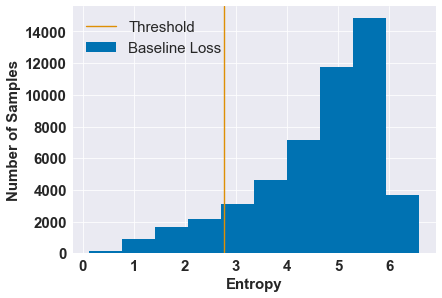}
    \caption{ImageNet Loss}
  \end{subfigure}
   \begin{subfigure}{0.24\linewidth}
    \includegraphics[width=\columnwidth]{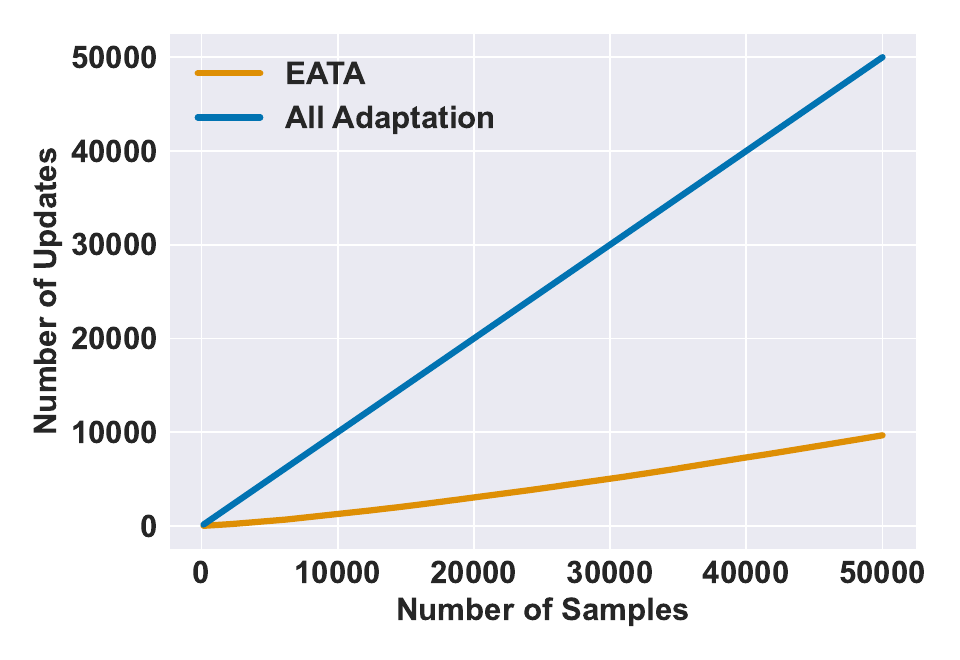}
    \caption{ImageNet EATA Updates}
  \end{subfigure}
  \caption{(a), (c): Entropy values prior to adaptation for different CIFAR-10 and ImageNet. (b), (d): Number of updates over adaptation for EATA.  All results are using the Gaussian noise corruption at severity 5.}
  \label{fig:num_updates}
\end{figure*}

%% file: experiments.tex
\section{Experiments}
\label{sec:results}

We show shortcomings of TTA with entropy minimization, and demonstrate benefits of REALM over existing approaches with batch size of one. We answer the questions: (1) Does a sample reliability criterion allow for sufficient sample updates?  (2) Does adapting on all samples irrespective of their reliability increase TTA performance? (3) How robust is REALM to model architecture, data quantity, and shift?
Additional details are in \cref{sec:exp_details}, and ablation studies are in \cref{sec:ablation}.
\subsection{Experimental Setup}

\noindent \textbf{Datasets and Models} We experiment with CIFAR-10-C and ImageNet-C benchmarks \cite{hendrycksbenchmarking}.  These datasets contain corrupted versions of the CIFAR-10 test set and ImageNet validation set according to four categories of corruptions (noise, blur, weather, and digital), for a total of 15 different corruptions.  We use the ResNet-26 model with group normalization following \cite{sun2020test} for CIFAR experiments, and a ResNet-50 with group normalization following \cite{niu2023towards} for ImageNet.

\noindent \textbf{Implementation Details} For CIFAR experiments, we use SGD with no momentum, and a batch size of one, unless otherwise stated.  For CIFAR-10 we set the learning rate to 0.005.  The initial values are $\alpha=0.15$ and $\lambda = 0.1$.  For ImageNet, we use SGD with momentum of $0.9$, a learning rate of 0.00025, and batch size of one. The initial $\alpha$ is the same, but $\lambda = 0.4\times \log(c)$ where  $c$ is the number of classes following \cite{niu2023towards}. The learning rate is scaled to account for small batch size following \cite{niu2023towards}. We also do not update model parameters for the last block of the network following \cite{niu2023towards}. For ImageNet, the loss starts large, and setting $C=\lambda$ offsets this impact on the gradient update.  For CIFAR-10, we set $C=1$. Additional hyperparameter and algorithm details are outlined in \cref{sec:exp_details}.

\begin{figure*}
  \centering
  \begin{subfigure}{0.24\linewidth}
    \includegraphics[width=\columnwidth]{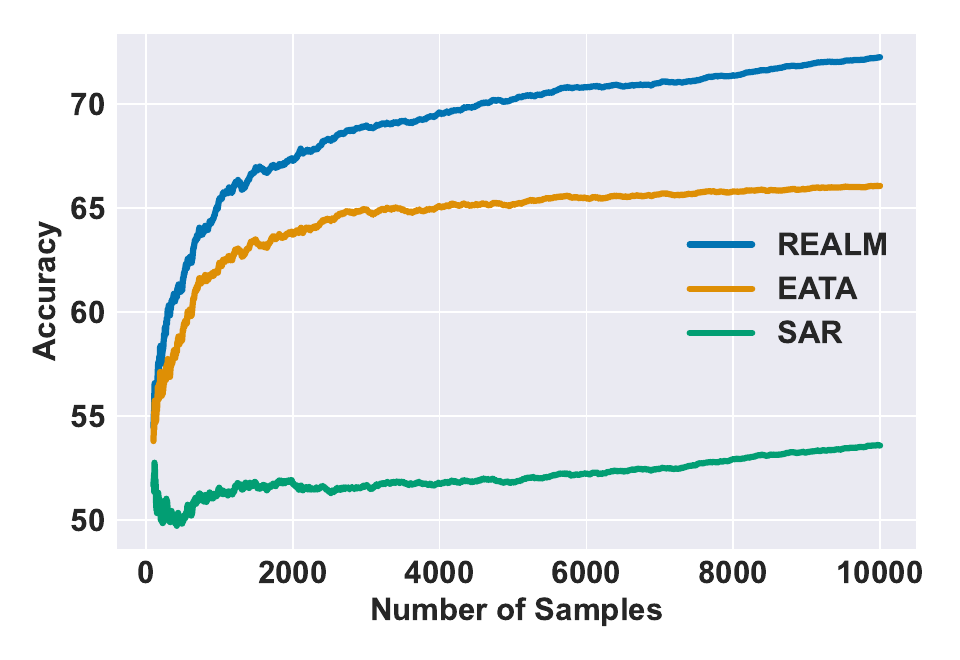}
    \caption{C10-GN}
  \end{subfigure}
  \hfill
  \begin{subfigure}{0.24\linewidth}
  \includegraphics[width=\columnwidth]{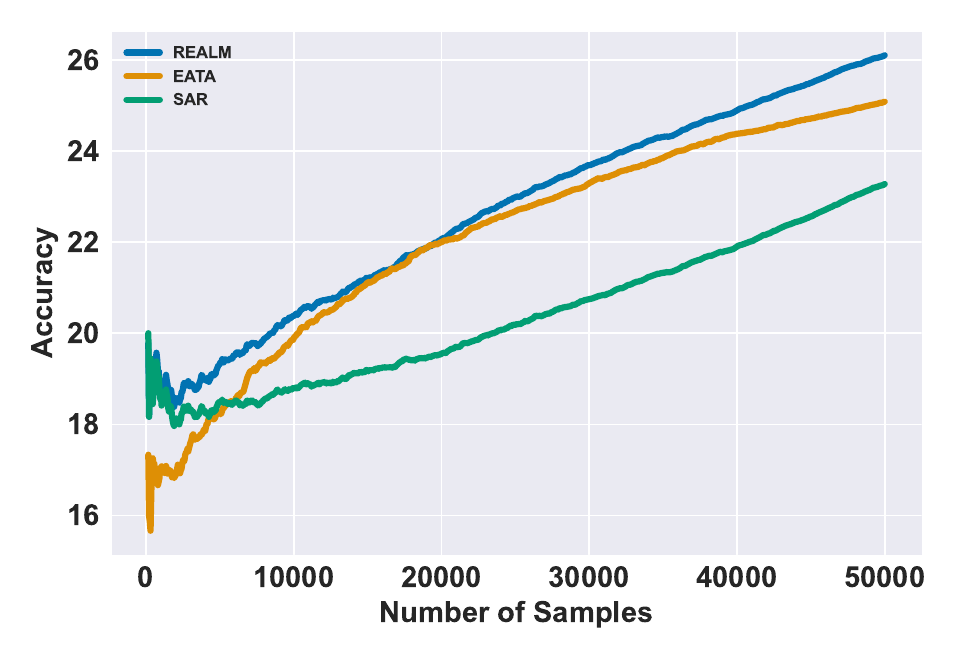}
    \caption{INet-GN}
  \end{subfigure}
  \begin{subfigure}{0.24\linewidth}
    \includegraphics[width=\columnwidth]{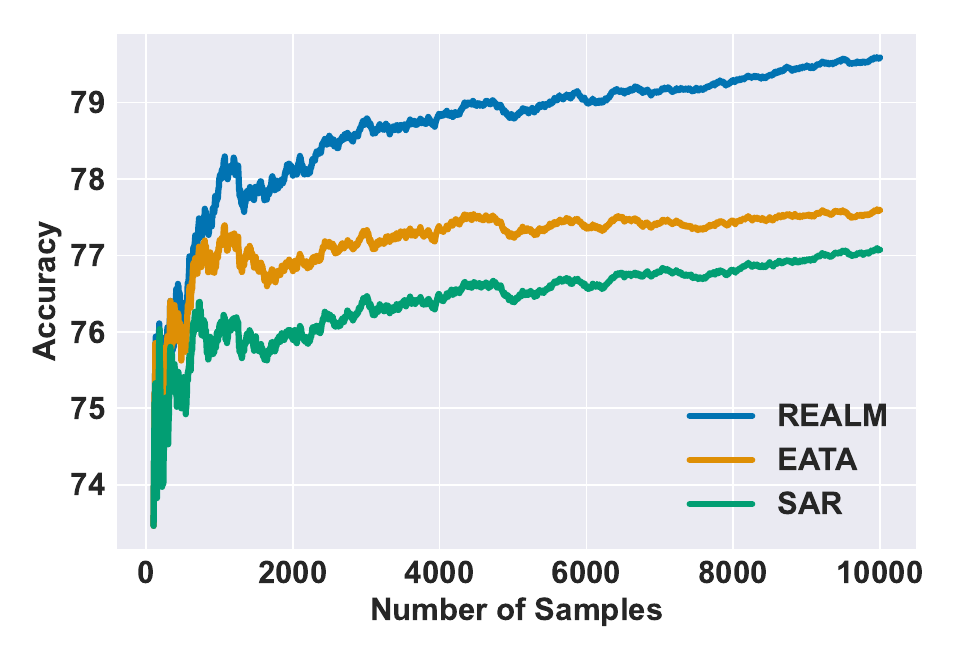}
    \caption{C10-Snow}
  \end{subfigure}
  \hfill
  \begin{subfigure}{0.24\linewidth}
  \includegraphics[width=\columnwidth]{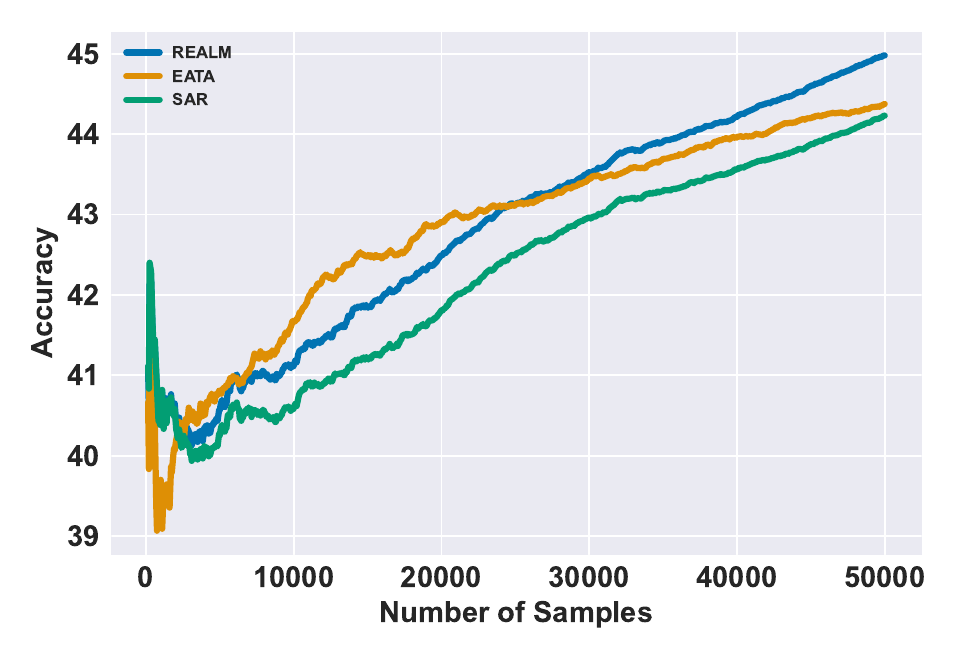}
    \caption{INet-Snow}
  \end{subfigure}
  \caption{Accuracy over adaptation according to the number of samples adapted on for the gaussian noise (GN) and snow corruptions at severity 5.  Accuracy is calculated as the fraction of samples correctly classified over all of the samples seen thus far. All runs use the same shuffling, and accuracy curves are averaged over 3 runs.}
  \label{fig:online_accuracy_over_samples}
\end{figure*}

\subsection{Limiting Sample Updates and Adaptation Frequency}
This section investigates the update frequency, and entropy of EATA, finding that EATA significantly limits the number of model updates, and has inconsistent updates across different loss distributions.  For the gaussian noise corruption at severity 5 (highest severity), we first plot the entropy of all samples on both the CIFAR-10 and ImageNet corrupted datasets in \cref{fig:num_updates}(a) and (c).  We find that the loss can be skewed in different ways, which can lead to differences in the update procedure.  For the CIFAR-10 model, the entropy is naturally low, with many samples below the EATA threshold, and so are used to adapt the model. In contrast, for ImageNet the losses are high initially, leading to slow adaptation at the start of training as the loss for the majority of samples is above the threshold.  It is important to note that depending on the order of the samples presented during TTA, this can lead to all samples for a particular class not being updated during adaption.  

To further highlight the impact of entropy thresholding, we plot the number of updates the model makes as a function of the number of samples the model has seen using the EATA method for adaptation. A low number of samples used implies that the model is not learning from the shifted distribution, and performance will remain relatively similar to the pre-adaptation.  Results are provided in \cref{fig:num_updates} (b) and (d), and show that the number of updates is less than two-thirds the number of samples for CIFAR-10.  This ratio is lower at the start of training where in the first $2000$ samples, only about one-half of the samples are used to adapt the model.  This makes the EATA weighting unreliable for a low number of adaptation steps.  On ImageNet, the number of samples updated drops to around 10,000 (only $20\%$ of the validation set).   


\subsection{Qualitative Results for Increasing Number of Adaptation Samples}
 We illustrated in the previous experiments that adaptation happens relatively slowly for EATA with only $20-50\%$ of the samples being used to update the model.  We now show this impacts the accuracy of online adaptation as each method progresses through test samples in CIFAR-C, and ImageNet-C in \cref{fig:online_accuracy_over_samples} for the gaussian noise and snow corruptions at severity 5.  
 
 We find that for CIFAR-10 gaussian noise, REALM achieves  higher accuracy consistently at the same number of samples during adaptation by $2-4\%$ indicating faster adaptation.  On ImageNet-C we note that performance is also higher at the start and end of adaptation by $1\%$. For the snow corruption, we note that performance increases consistently throughout adaptation for REALM.  On CIFAR-10 REALM starts to outperform EATA at around 1K samples, and outperforms at around 25K samples on ImageNet reaching a net gain of $1-2\%$.

\begin{table*}
    \centering
    \resizebox{\textwidth}{!}{
    \begin{tabular}{|l|ccc|cccc|cccc|cccc|>{\columncolor{LightCyan}}c|}
        \hline
         & \multicolumn{3}{c|}{Noise} & \multicolumn{4}{c|}{Blur} & \multicolumn{4}{c|}{Weather} & \multicolumn{4}{c|}{Digital} &  \\
         Method & Gauss. & Shot & Impul. & Defoc. & Glass & Motion & Zoom & Snow & Frost & Fog & Brit. & Contr. & Elastic & Pixel & JPEG & Avg.  \\\hline
         ResNet26 (GN)  & 51.6 & 55.2 & 49.7 & 75.9 & 52.3 & 75.5 &  75.9 & 75.9 & 66.9 & 72.0 & 85.9 & 70.3 & 74.4 & 56.3 & 71.7  & 67.3 \\
         + TTAug & 56.6 & 60.4 & 57.1 & 71.7 & 55.3 & 73.7 & 73.7 & 78.6 & 71.5 & 76.7 & 87.9 & 67.1 & 78.3 & 56.8 & \textbf{78.3} & 69.6\\
         + MEMO & 56.5 & 60.1 & 56.7 & 73.6 & 55.6 & 74.9 & 75.0 & 79.1 & 71.7 & 77.2 & \textbf{88.1} & 71.7 & \textbf{78.9} & 57.2 & \textbf{78.3} & 70.3\\
         + SFT & 68.1 & 73.3 & \textbf{71.1} & 82.3 & 55.8 & 81.6 & 79.8 & 79.2 & 76.6 & 79.3 & 86.1 & 74.6 & 75.5 & \textbf{78.1} & 74.9 & 75.8\\
         + EATA & 66.1 & 67.2 & 58.6 & 80.6 & 57.6 & 79.5 & 79.9 & 77.6 & 72.9 & 77.3 & 86.2 & 76.0 & 76.0 & 70.7 & 73.5 & 73.3\\
         + SAR & 53.6 & 59.1 & 49.9 & 79.8 & 53.0 & 78.3 & 79.2 & 77.1 & 70.5 & 76.1 & 86.2 & 74.5 & 75.7 & 60.5 & 72.6 & 69.7\\
         + REALM & \textbf{72.2} & \textbf{74.6} & 64.5 & \textbf{84.5} & \textbf{62.3} & \textbf{82.6} & \textbf{83.1} & \textbf{79.6} & \textbf{77.7} & \textbf{81.0} & 87.1 & \textbf{82.0} & 76.9 & \textbf{78.0} & 75.8 & \textbf{77.5}\\\hline    
    \end{tabular}
    }
    \caption{Accuracy across all corruptions in CIFAR10-C comparing REALM with prior SOTA methods. Results for REALM are averaged over 3 runs. }
    \label{tab:cifarc}
\end{table*}

\subsection{Comparison with SOTA Entropy Minimization Methods}
Table~\ref{tab:cifarc} summarizes results comparing REALM with many SOTA methods for online and episodic TTA for a ResNet-26 with Group Normalization layers (GN) at severity 5.  We find that on average, REALM performs the best across all corruptions, outperforming EATA, SAR, and Surgical Fine-Tuning (SFT), which trains specific convolutioanl layers of a network, and uses test-time augmentations based on augmix to create an ``artificial batch'' following \cite{zhang2022memo}.  REALM consistently performs as the top method across all corruptions except impulse noise, brightness, elastic transform, and JPEG. For all corruptions, except impulse noise, REALM is still the best performing method among the online adaptation methods. For impulse noise, REALM performs in the top two, performing a bit worse than SFT.  Results for MEMO, and TTAug are taken from \cite{zhang2022memo}, and results for SFT on gaussian noise and impulse noise corruptions are taken from \cite{lee2022surgical} as we found that training with the fixed lr from our hyperparameters resulted in the model predicting the same class on all inputs for a small number of our runs\footnote{In SFT, the authors perform a large hyperparameter experiment over lr, weight decay, and steps.  While our results are comparable to reported results for our choice of parameters, some small differences may cause instabilities we observed in our experiments.}.

\begin{table*}
    \centering
    \resizebox{\textwidth}{!}{
    \begin{tabular}{|l|ccc|cccc|cccc|cccc|>{\columncolor{LightCyan}}c|}
        \hline
         & \multicolumn{3}{c|}{Noise} & \multicolumn{4}{c|}{Blur} & \multicolumn{4}{c|}{Weather} & \multicolumn{4}{c|}{Digital} &  \\
         Method & Gauss. & Shot & Impul. & Defoc. & Glass & Motion & Zoom & Snow & Frost & Fog & Brit. & Contr. & Elastic & Pixel & JPEG & Avg.  \\\hline
         ResNet50 (GN)  & 18.0  & 19.8  & 17.9  & \textbf{19.8}  & 11.4  & 21.4  & 24.9  & 40.4  & \textbf{47.3}  & 33.6  & 69.3  & 36.3  & 18.6  & 28.4  & 52.3  & 30.6 \\
         + Tent       & 2.5  & 2.9  & 2.5  & 13.5  & 3.6  & 18.6  & 17.6  & 15.3  & 23.0  & 1.4  & 70.4  & 42.2  & 6.2  & \textbf{49.2}  & 53.8  & 21.5  \\
         + MEMO & 18.5  & 20.5  & 18.4  & 17.1  & 12.6  & 21.8  & 26.9  & 40.4  & \underline{47.0}  & 34.4  & 69.5  & 36.5  & 19.2  & 32.1  & 53.3  & 31.2  \\ 
         + DDA       & \textbf{42.4} & \textbf{43.3} & \textbf{42.3} & 16.6 & \textbf{19.6} & 21.9 & 26.0 & 35.7 & 40.1 & 13.7 & 61.2 & 25.2 & \textbf{37.5} & 46.6 & 54.1 & 35.1 \\
         + EATA       & 24.8  & 28.3  & 25.7  & 18.1  & 17.3  & 28.5  & 29.3  & 44.5  & 44.3  & \underline{41.6}  & 70.9  & \textbf{44.6}  & 27.0  & 46.8  & \underline{55.7}  & 36.5  \\
         + SAR  & 23.4 & 26.6 & 23.9 & \underline{18.4} & 15.4 & \underline{28.6} & \textbf{30.4} & \textbf{44.9} & 44.7 & 25.7 & \textbf{72.3} & \underline{44.5} & 14.8 & 47.0 & \textbf{56.1} & 34.5 \\
         + REALM &  \underline{26.9} & \underline{29.9} & \underline{28.0} & \underline{18.4} & \underline{18.2} & \textbf{29.6} & \textbf{31.1} & \textbf{45.6} & 43.6 & \textbf{45.5} & \underline{71.2} & 44.4 & \underline{28.9} & \textbf{49.7} & 55.5 & \textbf{37.8} \\\hline
    \end{tabular}
    }
    \caption{Accuracy across all corruptions in ImageNet-C comparing REALM with prior SOTA. Results are averaged over 3 runs.}
    \label{tab:inetc}
\end{table*}

Table~\ref{tab:inetc} summarizes comparison of REALM with SOTA TTA methods for a ResNet-50 GN at severity 5. Results are taken from \cite{niu2023towards}, however we confirmed performance for both EATA, and SAR.  REALM performs the best on average across all corruptions, outperforming EATA, SAR, and DDA (which uses a diffusion model trained on the same in-distribution set).  REALM is consistently a top two method across all corruptions except for frost, contrast, and jpeg compression.  In these corruptions, REALM is competitive with the top performing method.  

\subsection{Comparisons with Different Model Architectures}
We experiment with classifiers beyond the ResNet models used in the previous section to evaluate whether REALM's improvements apply to arbitrary architectures.  Table~\ref{tab:mult_archs} shows results for four different architectures on the gaussian noise corruption at severity 5 in ImageNet-C.  Following \cite{gao2023back}, we select ResNet-50 GN, two transformer architectures: VitBase-LN (vision transformer with layer norm), Swin-tiny transformer, and a ConvNext-tiny network as these are all state-of-the-art attentional and convolutional architectures with roughly the same number of parameters.  The Vitbase model is trained with a learning rate of $0.001/64$ following \cite{niu2023towards}.  All other  models are trained with the same hyperparameters as the ResNet architecture.

\begin{table}[ht]
    {\small
    \centering
    \begin{tabular}{|l|cccc|>{\columncolor{LightCyan}}c|}
    \hline
    Method & RNet-50 & ViT & SWIN & ConvNext & Avg. \\\hline
    EATA & 24.8 & 31.4 & \textbf{38.5} & 48.5 & 35.8\\
    SAR & 23.3 & \textbf{41.0} & 31.3 & 51.4 & 36.8\\
    REALM & \textbf{26.1} & 36.4 & 36.0 & \textbf{54.8} & \textbf{38.3} \\
    \hline
    \end{tabular}
    \caption{Accuracy over four different networks: ResNet-50 with group normalization layers, ViT-base, Swin-tiny, and ConvNext-tiny pretrained on ImageNet-1k.  Comparisons are done on Imagenet-C gaussian noise corruption with severity 5 and results are averaged over 3 runs.}
    \label{tab:mult_archs}
    }
\end{table}

For both convolutional networks, REALM performs the best, while on the transformer models, REALM performs slightly worse.  Both EATA and SAR perform inconsistently across the architectures with instances where they perform almost 10\% worse than the best performing method.  Nonetheless, we believe that no method consistently outperforms across architectures, and investigating architectural differences with TTA methods should be the subject of future work.

\subsection{Comparison with Few Adaptation Samples}
We experiment with adaptation of the ResNet model used in the previous section to evaluate whether REALM's improvements apply under limited sample settings.  Table~\ref{tab:few_samples} shows results for increasing total number of samples for adaptation on the Gaussian noise corruption at severity 5 in ImageNet-C.  To evaluate models adapted on a subset of the data, we holdout the last 10k samples from the test set.  All  models are trained with the same hyperparameters.  

\begin{table}[h!]
    {\small
    \centering
    \begin{tabular}{|l|ccccc|}
    \hline
    Method & 1024 & 2048 & 4096 & 10k & 20k\\\hline
    ResNet  & 17.7 & 17.7 & 17.7 & 17.7 & 17.7\\
    + EATA &  \textbf{18.2} & 18.3 & 19.0 & 21.4 & 24.6\\
    + SAR &  17.8 & 17.9 & 18.0 & 19.2 & 21.4\\
    + REALM  & 18.1 & \textbf{18.6} & \textbf{19.5} & \textbf{21.6} & \textbf{25.2}\\
    \hline
    \end{tabular}
    \caption{Results for increasing number of adaptation samples for a ResNet-50 GN. Comparisons are done on Imagenet-C gaussian noise corruption with severity 5 and results are averaged over 3 runs.}
    \label{tab:few_samples}
    }
\end{table}

We find that REALM outperforms EATA starting from 2048 adaptation samples by around 0.5\%, and outperforms SAR at all number of adaptation samples.  We also note that accuracy increases consistently with increasing number of adaptation samples indicating better domain generalization capability with additional samples. Finally, we see that all methods have the largest jump in improvement at 10K and REALM reaches similar performance on the held-out set after 20k adaptation steps that EATA and SAR reach adaptation on the full validation set.  

\subsection{Comparisons with Different Datasets}

We further conduct experiments on additional ImageNet distribution shift datasets: ImageNet-Renditions (R) \cite{hendrycks2021many} and ImageNet-Adversarial (A) \cite{Hendrycks_2021_CVPR}.  Differing from corruption robustness in ImageNet-C, ImageNet-R and ImageNet-A contain real samples that are difficult for models trained without domain generalization properties to classify.  In particular, ImageNet-R contains renditions of a subset of the classes in ImageNet including paintings, sculptures, embroidery, cartoons, origami, and toys. Imagenet-A contains images collected from iNaturalist and Flickr that are incorrectly classified by a ResNet-50. Additional details about the dataset are available in \cref{sec:exp_details}.

\begin{table}[ht]
    {\small
    \centering
    \begin{tabular}{|l|cc|>{\columncolor{LightCyan}}c|}
    \hline
    Method & ImageNet-R & ImageNet-A \\\hline
    No Adapt & 40.8 & 0.1\\
    REALM & \textbf{42.5} & \textbf{14.3} \\
    \hline
    \end{tabular}
    \caption{Results for ResNet-50 GN evaluated on the ImageNet-R and ImageNet-A datasets.}
    \label{tab:inet_other}
    }
\end{table}

Results on these datasets for REALM are shown in \cref{tab:inet_other} indicating REALM improves  performance on both datasets over no adaptation, and that REALM improves performance on distribution shifts outside common corruptions.


%% file: conclusion.tex
\section{Conclusion}
\label{sec:conc}

This work illustrates the shortcomings of TTA in the online single instance batch setting. We highlight that current approaches aimed at stabilizing TTA by skipping unreliable samples with high entropy, result in no adaptation on a large portion of the dataset. We then show equivalence to SPL with a specific regularizer, and introduce REALM our  approach for stabilizing online TTA entropy minimization approaches. REALM improves on prior approaches by penalizing the update of all samples using a robust function of the entropy rather than skipping the sample based on entropy entirely. This yields improved results on corruptions of CIFAR-10 and ImageNet.  Additionally, REALM is simple to implement, requiring only modification of the loss function, and is theoretically grounded within the framework of self-paced learning.  We believe this work is a step towards creating more robust models through online TTA.

%% file: appendix2.tex
\appendix
\begin{section}{REALM Details}
\label{sec:realm_details}
In this section, we detail the derivation showing that REALM is an SPL objective, and provide pseudocode for our implementation of REALM in \cref{algorithm:realm}. 
\subsection{REALM as an SPL Objective}
\label{sec:realm-robust-regularized}
As outlined in \cref{sec:realm}, the EATA procedure can be re-cast as a SPL method with the explicit regularizer 
\begin{equation}
  g(w; \lambda) = -\lambda\|w\|_1. 
\end{equation}

Similarly, minimizing the loss function
\begin{equation}
    \rho(x; \alpha, \lambda) = \frac{|\alpha-2|}{\alpha}\cdot  C \cdot \left[\left(\frac{(x/\lambda)}{|\alpha-2|}+1\right)^{\alpha/2}-1\right], 
    \label{eqn:rho_appendix}
\end{equation}
 used in our REALM framework can be reinterpreted as solving a regularized optimization problem similar to
\begin{equation}
\begin{split}
    w^*, \theta^* &= \argmin_{w, \theta}\mathbbm{E}\left[w(x)\mathcal{L}(\theta; x) + g(w; \lambda)\right],\\
    &= \argmin_{w, \theta}\frac{1}{N}\sum_{i=1}^N \left[w(x_i)\mathcal{L}(\theta; x_i) + g(w(x_i); \lambda)\right],
\end{split}
\label{eqn:regularized-min-appendix}
\end{equation}
with a specific regularizer $g(w;\alpha, \lambda)$. In this section, we aim to derive an explicit formula for $g(w;\alpha, \lambda)$.

For our analysis, we closely follow the derivation in \cite{unification1996}. The goal is to equivalently cast the robust minimization problem REALM:
\begin{equation}
    \theta^*, \alpha^*, \lambda^* = \min_{\theta, \alpha, \lambda} S_{div}(x) \rho\left(\mathcal{L}(\theta; x); \alpha, \lambda\right), 
    \label{eqn:robust-min-appendix}
\end{equation}
which we simplify as
\begin{equation}
	\min_{\theta} \rho\left(\mathcal{L}(\theta);\alpha,\lambda\right),
	\label{eqn:robust-min-simple}
\end{equation}
as a regularized minimization problem, in the form
\begin{equation}
	\min_{\theta,w\in\left[0,1\right]}\left[w\frac{\mathcal{L}(\theta)}{\lambda} + g(w;\alpha)\right].
	\label{eqn:regularized-min-simple}
\end{equation}
Note that in \cref{eqn:robust-min-simple} we are ignoring the term $S_{div}$, which is treated as a constant; we are also not considering the optimization with respect to $\alpha$ and $\lambda$, since this can be treated separately. Moreover, with a slight abuse of notation, let us re-define more compactly $t=\mathcal{L}(\theta)/\lambda$ as the argument of $\rho$, and not report the dependency on $\alpha$ explicitly, that is
\begin{equation}
    \rho(t)\coloneqq \frac{|\alpha-2|}{\alpha}\left(\left( \frac{2t}{|\alpha-2|} + 1 \right)^{\frac{\alpha}{2}}-1\right).
\end{equation}

For the equivalence between \cref{eqn:robust-min-simple} and \cref{eqn:regularized-min-simple} to hold, we must ensure that minimizing either term over $\mathcal{L}(\theta)$ results in the same solution. To this end, we impose the equivalence of both
\begin{equation}
	\rho\left(t\right) = \min_{w}\left[wt + g(w)\right],
	\label{eqn:cond1}
\end{equation}
as well as their derivatives with respect to $\mathcal{L}(\theta)$,
\begin{equation}
	\frac{\rho'\left(t\right)}{\lambda} = \frac{w}{\lambda} \quad\Longrightarrow\quad \rho'(t) = w.
	\label{eqn:cond2}
\end{equation}
Differentiating \cref{eqn:cond1} with respect to $w$ and substituting \cref{eqn:cond2}, at the optimal point we get
\begin{equation}
	t + g'\left(\rho'(t)\right) = 0.
    \label{eqn::ode-g}
\end{equation}
Multiplying by $\rho''(t)$ and integrating by parts, we recover
\begin{equation}
\begin{split}
	&g'\left(\rho'(t)\right)\rho''(t) = -t\rho''\left(t\right)\\
	\Longleftrightarrow& \left(g\left(\rho'(t)\right)\right)' = -\left(t\rho'(t)\right)' + \rho'(t)\\
	\Longleftrightarrow& g\left(\rho'(t)\right) = -t\rho'(t) + \rho(t)\\
	\Longleftrightarrow& g\left(w\right) = -w \left(\rho'\right)^{-1}(w) + \rho\left(\left(\rho'\right)^{-1}(w)\right),
\end{split}
\label{eqn:regularizer-implicit}
\end{equation}
where again we substituted \cref{eqn:cond2} to express the regularizing term $g(w)$ as a function of $w$. Notice that indeed the inverse of $\rho'(t)$ is well-defined for our choice of $\rho(t)$: we have in fact, after some simplifications,
\begin{equation}
\begin{split}
	&\rho'(t) = \left(\frac{2t}{|\alpha-2|}+1\right)^{\frac{\alpha-2}{2}}\\
	\Longrightarrow&\left(\rho'\right)^{-1}(w) = \frac{|\alpha-2|}{2}\left(w^{\frac{2}{\alpha-2}}-1\right)\\
\end{split}
\label{eqn:rho-prime_appendix}
\end{equation}
Substituting this into \cref{eqn:regularizer-implicit} allows us to write $g(w)$ more explicitly:
\begin{equation}
	g(w;\alpha) = \frac{|\alpha-2|}{\alpha}\left(w^{\frac{\alpha}{\alpha-2}}\left(1-\frac{\alpha}{2}\right) + \frac{\alpha}{2} w - 1\right).
\label{eqn:regularizer-explicit}
\end{equation}

Now that we have a candidate form for $g(w;\alpha, \lambda)$, we need to verify that this indeed identifies a valid regularizer. 
First, $w=\phi'(t)$ must be an actual minimum for \cref{eqn:regularized-min-simple}: in other words, we must have
\begin{equation}
	\frac{\partial^2}{\partial w^2}\left(w\frac{\mathcal{L}(\theta)}{\lambda} + g(w;\alpha)\right)>0 \quad\Longrightarrow\quad g''(w;\alpha)>0.
\end{equation}
This condition can be equivalently rewritten by taking the derivative of \cref{eqn::ode-g} with respect to $t$,
\begin{equation}
    \left(t+g'(\rho'(t))\right)' = 0 \quad\Longrightarrow\quad g''(\rho'(t)) = -\frac{1}{\rho''(t)},
\end{equation}
which shows that, for $g(w,;\alpha)$ to be convex, it suffices to ask for $\rho(t)$ to be concave. This can be promptly verified:
\begin{equation}
    \rho''(t) = \frac{\alpha-2}{|\alpha-2|}\left(\frac{2t}{|\alpha-2|}+1\right)^{\frac{\alpha}{2}-2}< 0\quad\text{for }\alpha<2.
    \label{eqn:phi-prime-decreasing}
\end{equation}
Incidentally, this also implies $\rho''(t)\neq0$, which we made use of in our derivation; moreover, in light of this, $\rho'(t)$ is monotone decreasing. We also need $w=\rho'(t)$ to span the whole domain of $w\in[0,1]$, namely
\begin{equation}
    \lim_{t\to0}\rho'(t)=1, \quad\text{and}\quad\lim_{t\to\infty}\rho'(t)=0.
    \label{eqn:phi-prime-bounded}
\end{equation}
This too can be verified from its definition in \cref{eqn:rho-prime_appendix}, and holds for $\alpha<2$. Finally, notice that \cref{eqn:phi-prime-decreasing,eqn:phi-prime-bounded} above correspond to the conditions in \cite[Definition 1]{kumar2010self}, further confirming that the REALM robust loss function falls within the SPL framework.

\paragraph{A note on the adaptation of $\rho$ for entropy minimization}
The original definition of the robust loss function appearing in \cite{barron2019general} reads:
\begin{equation}
    \rho^{0}(\mathcal{L};\alpha,\lambda)\coloneqq
    \frac{|\alpha-2|}{\alpha} \left(\left(\frac{(\mathcal{L} / \lambda)^2}{|\alpha-2|} + 1\right)^{\alpha/2} - 1\right).
    \label{eqn:rhoOriginal}
\end{equation}
Notice that, compared against our definition in \cref{eqn:rho_appendix}, the argument of this function $(\mathcal{L}/\lambda)$ appears squared: in fact, \cref{eqn:rhoOriginal} was originally intended for a squared-error type of loss function. Indeed, starting from the original definition of $\rho^{0}$ and following a similar derivation as the one outlined in \cref{sec:realm-robust-regularized}, one can show that the corresponding regularized minimization problem is given by
\begin{equation}
	\min_{\theta,w\in\left[0,1\right]}\left[\frac{1}{2}w\left(\frac{\mathcal{L}(\theta)}{\lambda}\right)^2 + g(w;\alpha)\right],
	\label{eqn:regularized-min-original}
\end{equation}
rather than \cref{eqn:regularized-min-simple}: that is, the objective of the regularized problem is alsosquared. Since our target loss function is entropy (rather than squared entropy), we adapted $\rho$ accordingly in \cref{eqn:rho_appendix}, so to ensure consistency between the objectives of the robust and regularized minimization problems.

\subsection{Pseudocode for REALM}
Pseudocode for REALM is given in Algorithm~\ref{algorithm:realm}.  REALM is relatively easy to implement requiring only additional computation of the robust loss function to scale the entropy, and gradient updates for both $\alpha$, and $\lambda$.  This is the objective used in REALM as ut satisfies the desired properties, and results in a scaled entropy objective as desired.

\begin{algorithm*}[ht]
  \linespread{1.25}\selectfont
  \caption{\textsc{REALM}: Robust Entropy Adaptive Loss Minimization}
  \label{algorithm:realm}
\begin{algorithmic}[1]
   \STATE {\bfseries Input:} $\mathcal D_{\textrm{Test}}$, $f(\cdot \, ; \, \theta_0)$, $\alpha_0$, $\lambda_0$\\[3pt]
   \FOR{$t=0$ {\bfseries to} $T$}
        \STATE Compute predictions $\hat{y}_t = f(x_t; \theta)$
        \STATE Compute robust loss $\mathcal{L}^* = \rho(\mathcal{L}(\hat{y}_t))$
        \STATE Compute weight $S_{\textrm{div}} = \mathbbm{1}\left\{\cos\left(f(x_t; \theta), m_{t-1}\right)\ < d \right\}$
        \STATE Scale the robust loss $\mathcal{L}^* \leftarrow S_{\textrm{div}}\mathcal{L}^*$
        \STATE Update $\theta_{t+1} \leftarrow \theta_{t} - \eta\nabla_{\theta} \mathcal{L}^*$
        \STATE Update $\alpha_{t+1}, \lambda_{t+1} \leftarrow \alpha_{t}, \lambda_{t} - \eta\nabla_{\alpha,\lambda} \mathcal{L}^*$
   \ENDFOR
\end{algorithmic}
\end{algorithm*}

\subsection{Sources of Entropy Collapse}
Methods such as Tent and MEMO lead to model collapse resulting in all samples having the same class prediction due to online entropy optimization with single sample batch sizes.  While a primary cause of this collapse is noisy samples, which we aim to handle in this work, there may be other causes including high learning rates \cite{choi2022improving}, or mixed distribution shifts \cite{niu2023towards}. These sources of collapse are outside the scope of our work.

\end{section}

\newpage

\section{Experimental Details for REALM}
\label{sec:exp_details}
In this section, we provide additional details for models, datasets, and hyperparameters for all methods.

\subsection{Additional Model Details}
\noindent \textbf{ResNet-26 GN} - We use the ResNet-26 network from \cite{sun2020test}.  The model is trained on the CIFAR-10 train set.  

\noindent \textbf{ResNet-50 GN} - We use the ResNet-50 GN architecture with pretrained weights from the {\tt timm} library available under the name {\tt resnet50\_gn}. The model is trained on the ImageNet train set.  

\noindent \textbf{Vit} - We use the Vit base architecture with pretrained weights from the {\tt timm} library available as {\tt vit\_base\_patch16\_224}.  The model is trained on the ImageNet train set.

\noindent \textbf{Swin} - We use the Swin tiny architecture with pretrained weights from {\tt timm} available as {\tt swin\_tiny\_patch4\_window7\_224}.  The model is trained on the ImageNet train set.

\noindent \textbf{ConvNext} - We use the ConvNext tiny architecture with pretrained weights from the {\tt timm} library as {\tt convnext\_tiny}.  The model is trained on ImageNet train set.

Both the ResNet and Vit models are evaluated for comparison to \cite{niu2023towards}.  The Swin and ConvNext architectures are SOTA models used to demonstrate that REALM performs well across architectures.  We use the tiny versions to maintain similar parameter counts to the ResNet-50. We note that variability in performance is likely not a result of the number of model parameters, rather architectural differences such as attention layers.

\subsection{Additional Dataset Details}

\noindent \textbf{CIFAR-10-C} - CIFAR-10-C is a collection of 15 different corruptions types from four categories (noise, blur, weather, and digital) applied to the CIFAR-10 test set across five different severity levels.  We evaluate all approaches on the highest severity.  Each corruption has a total of 10,000 samples matching the CIFAR-10 test set \cite{hendrycksbenchmarking}.

\noindent \textbf{ImageNet-C} - ImageNet-C is a collection of 15 different corruptions types from four categories (noise, blur, weather, and digital) applied to the ImageNet validation set across five different severity levels.  We evaluate all approaches on the highest severity.  Each corruption has a total of 50,000 samples matching the original validation set \cite{hendrycksbenchmarking}.

\noindent \textbf{ImageNet-R} - ImageNet-R contains renditions of a subset of the classes in ImageNet including paintings, sculptures, embroidery, cartoons, origami, and toys.  A total of 30,000 samples  across 200 of the ImageNet classes are contained in ImageNet-R.  For TTA on ImageNet-R we  subset the network outputs before adapting \cite{hendrycks2021many}.  

\noindent \textbf{ImageNet-A} - ImageNet-A contains samples collected from Flickr and iNaturalist according to a subset of 200 of the classes in ImageNet.  All samples collected are incorrectly classified by a ResNet model, and the probability of the correct class is lower than $15\%$. A total of 7,500 adversarially filtered samples are used for adaptation.  For TTA on ImageNet-A we  subset the network outputs before adapting \cite{Hendrycks_2021_CVPR}.

\subsection{Additional Hyerparameter Details}
We outline hyperparameters for all methods.  For all models, we adapt only the normalization layer parameters following \cite{wangtent}.

\noindent \textbf{TENT} - For CIFAR-10 we use SGD with no momentum, and a batch size of one. We set the learning rate to 0.005. For ImageNet, we use SGD with momentum of $0.9$, a learning rate of 0.00025, and batch size of one. The learning rate is scaled to account for small batch size as $\textrm{lr} = (\textrm{lr} / 32)$ following \cite{niu2023towards}. The initial learning rate for the Vit model is set to $0.001$ and is scaled similarly.

\noindent \textbf{EATA} - In addition to the hyperparameters used in TENT, we set $\lambda = 0.4\times \log(c)$ where $c$ is the number of classes in the dataset.  For CIFAR-10, the threshold for $S_{\textrm{div}}$ is set to $0.4$.  For ImageNet, the threshold is set to $0.05$. 

\noindent \textbf{SAR} - In addition to the hyperparameters for EATA and TENT, we scale the learning rate as  $\textrm{lr} = (\textrm{lr} / 16)$ except for the Vit model, and we freeze the last block of the network.  For CIFAR-10, we use a much smaller threshold at $\lambda = 0.1$ as the loss is much smaller than on ImageNet, and we found that adapting on samples with $\mathcal{L} \in [0.1, 0.4\times \log(10)]$ resulted in unstable adaptation.  

\noindent \textbf{SFT} - We follow the same hyperparameters as those for TENT, except we adapt only the first conv layer of the network.  We did not scale the learning rate as the method creates a batch of data via augmentations. We set the number of augmentations to 64 following the batch size used in EATA \cite{niu2022efficient}.  For MEMO, we use an identical implementation, only we do not freeze any part of the network.

\noindent \textbf{REALM} - We follow the same hyperparameters as SAR for fair comparison.  We set the initial $\alpha=0.15$, $\lambda=0.1$ for CIFAR-10, and $\lambda=0.4\times \log(1000)$ for ImageNet.  We did not tune these values for fair comparison to SAR.  The value of $\alpha$ is chosen as it is close to $0$ and mimics the behavior of the reliable sample criteria.  We set the learning rate for $\alpha$ and $\lambda$ to a factor of 2 of the model parameter learning rates for CIFAR-10 and the same for ImageNet.  Note that in \cref{sec:ablation} we find that setting the learning rate to the same  as that for the model parameters, and dropping the learning rate improves performance.  However, extra tuning of the hyperparameters outside using the same learning rate as for the model parameters may give an unfair advantage to REALM over competing methods. For fair comparison across methods in the main text we did not tune these.    

\section{Additional Commentary on REALM in Single Sample TTA}
\label{sec:add_commentary}
\subsection{Results on Forgetting}

Our primary focus in the main paper is to demonstrate that REALM improves online adaptation performance generalizing better to the target distribution.  However, an unwanted consequence of adaptation is forgetting the original distribution resulting in decreased performance on the original in-distribution data.  We investigate whether REALM results in additional forgetting over a method such as EATA \cite{niu2022efficient}, which reduces forgetting explicitly via the anti-forgetting regularizer in \cite{kirkpatrick2017overcoming}.  To investigate, we adapt the model on the ImageNet-C gaussian noise corruption at severity 5, then re-evaluate model performance on the original validation set. Results are summarized in \cref{tab:forgetting} indicating that adaptation with REALM results in little to no drop in performance on the clean validation data.

\begin{table}[ht]
    {\small
    \centering
    \begin{tabular}{|l|cc|}
    \hline
    Method & Gaussian Noise & Clean \\\hline
    No Adapt & 18.0 & \textbf{80.0}\\
    EATA & 24.9 & 79.8\\
    REALM & \textbf{26.1} & 79.6 \\
    \hline
    \end{tabular}
    \caption{Accuracy for ResNet-50 GN evaluated after adaptation on Gaussian noise corruptions and clean validation set. Results are averaged over 3 runs.}
    \label{tab:forgetting}
    }
\end{table}

We did additional experiments using the anti-forgetting regularizer in \cite{kirkpatrick2017overcoming}. Using the same hyperparameters as in \cite{niu2022efficient}, we found that we were able to precisely maintain performance on in-distribution data, however dropped performance on the gaussian noise corruption indicating that there is a tradeoff in choosing the correct hyperparameter to control the regularizer.  For other datasets outside the ones reported in this paper, it is possible that adding the additional regularizer is necessary to preserve performance on in-distribution data, however, for our experiments, we found that performance was already comparable on clean data without the need for the additional regularizer. 

\subsection{Wall-Clock Runtime Comparison}
We implement REALM, SAR, and EATA using the Pytorch library \cite{paszke2017automatic}, and report the wall-clock time adaptation takes for a single sample on average\footnote{We use the torch profiler and time the {\tt model\_inference} component according to the example in \url{https://pytorch.org/tutorials/recipes/recipes/profiler_recipe.html}}.  For the ResNet-50 with group normalization layers on ImageNet, all methods with no gradient update  take between 5-10ms with standard inference taking 5ms, REALM taking 7ms, and EATA taking 10ms.  When the model backwards on the single instance, REALM finishes the fastest at around 45ms, EATA takes 60ms, and SAR takes 80ms.  Further note that in the case of batch updates, although fewer samples may be needed for the backward pass, a standard implementation cannot take advantage of this, and will result in similar wall-clock time per batch as found in \cite{niu2022efficient}.  Thus, without specialized hardware, in practical experiments, we've found REALM's runtime to be comparable to EATA and faster than SAR.

\subsection{Additional Commentary on Training Robustness}
Prior studies investigate methods for improving robustness to distribution shifts during training.  A common practice is to train with augmentations aimed at smoothing the training data distribution, and improving robustness to covariate shifts.  Some examples include Augmix \cite{hendrycksaugmix}, AutoAugment \cite{cubuk2019autoaugment}, Adversarial augment \cite{zhangadversarial}, and Mixup \cite{zhangmixup}.  This approach for attaining robustness happens at training time, and our TTA approach is agnostic to this. Further, improving robustness at training time is difficult as one needs to carefully construct an augmentation policy which covers any expected distribution shifts.  In contrast, applying such augmentations at test-time as is done in prior works \cite{ lee2022surgical, liu2021ttt++,zhang2022memo} does not yield better performance overall, and makes the adaptation procedure much slower. 

\subsection{Additional Commentary on Batch Size of One}
Our focus in this work is on the setting where online TTA is performed with a batch size of one.  Recall our example where an individual is taking picture(s) on their phone on a rainy day. A phone app providing object recognition capabilities would be expected to immediately classify on the new sample. We reiterate that this is an important setting, as it may be infeasible to wait for enough samples to batch, and one may not see enough samples to form a batch.  Further, in the batch size of one setting, batch normalization can fail to adapt parameters and running statistics on a single sample from the new distribution, and methods that rely on sample skipping can skip a large portion of the test samples. In this work, we do not explore larger batch size setting as we believe that online inference is less practical when waiting for a batch of data, especially when data aggregation is not feasible due to privacy or storage concerns, and  optimizing over a subset of the batch has little practical advantage in terms of efficiency as long as the original batch fits in memory without specialized software, or hardware.

\begin{table*}
    \centering
    \resizebox{\textwidth}{!}{
    \begin{tabular}{|l|ccc|cccc|cccc|cccc|>{\columncolor{LightCyan}}c|}
        \hline
         & \multicolumn{3}{c|}{Noise} & \multicolumn{4}{c|}{Blur} & \multicolumn{4}{c|}{Weather} & \multicolumn{4}{c|}{Digital} &  \\
         Method & Gauss. & Shot & Impul. & Defoc. & Glass & Motion & Zoom & Snow & Frost & Fog & Brit. & Contr. & Elastic & Pixel & JPEG & Avg.  \\\hline
         ResNet50 (GN)  & 18.0  & 19.8  & 17.9  & \textbf{19.8}  & 11.4  & 21.4  & 24.9  & 40.4  & \textbf{47.3}  & 33.6  & 69.3  & 36.3  & 18.6  & 28.4  & 52.3  & 30.6 \\
         + REALM ($2.5e-4$) & 26.9 & 29.9 & 28.0 & 18.4 & 18.2 & 29.6 & 31.1 & 45.6 & 43.6 & 45.5 & 71.2 & 44.4 & 28.9 & 49.7 & 55.5 & 37.8\\
         + REALM ($5e-4$)& 26.1 & 28.9 & 27.2 & 18.3 & 17.7 & 29.1 & 30.4 & 45.0 & 43.2 & 44.8 & 71.0 & 43.9 &   27.9 & 49.1 & 55.3 & 37.2 \\
          + REALM ($5e-7$) & \textbf{27.9} & \textbf{31.1} & \textbf{29.2} & 18.6 & \textbf{18.9} & \textbf{30.4} & \textbf{32.1} & \textbf{46.4} & \textbf{44.1} & \textbf{46.3} & \textbf{71.5} & \textbf{45.1} & \textbf{30.1} & \textbf{50.4} & \textbf{55.8} & \textbf{38.5} \\\hline
    \end{tabular}
    }
    \caption{Accuracy across all corruptions in ImageNet-C comparing REALM with different learning rates for $\alpha$ and $\lambda$. Results are averaged over 3 runs.}
    \label{tab:small_lr}
\end{table*}

\section{Ablation Studies}
\label{sec:ablation}
REALM reduces the impact of outliers during online adaptation by applying the robust loss $\rho$ to penalize the gradient update of high entropy samples.  While our approach is theoretically grounded in the self-paced learning literature, its implementation is rather straightforward.  In this section, we study the impact of several parameter choices, and implementation details of our objective.  In particular, we examine the impact of $S_{\textrm{div}}$,  different terms in the loss function, reduced gradient updates on all model blocks, and different initial values for the loss. For all ablations, we report results for the gaussian noise corruption at severity 5, however we expect results to be consistent across all corruptions as removing components, and changing the objective will harm performance and nullify theoretical justification. In some cases, when tuning hyperparamters unique to REALM, we show that improved performance can be achieved.  In particular, tuning the learning rate for $\alpha$ and $\lambda$ and updating the last block of the network are both shown to increase corruption robustness.  In the main text, however, we do not tune these extra hyperparameters to keep as fair comparison as possible between REALM and competing methods.  Nonetheless, if one is able to tune these extra parameters, higher performance can be achieved.

\subsection{On the Importance of $S_{\textrm{div}}$}
Recall that for the EATA adaptation procedure there are two weights: (1) $S_{\textrm{ent}}$, which controls adaptation on outliers based on their entropy, and  (2) $S_{\textrm{div}}$, which controls adaptation redundancy based on whether the predictions are similar.  We connect $S_{\textrm{ent}}$ to optimization of the self-paced learning objective, and treat   $S_{\textrm{div}}$ as independent as no gradient is applied through this weight.  Results are summarized in \cref{tab:sdiv}.  

\begin{table}[ht]
    {\small
    \centering
    \begin{tabular}{|l|c|}
    \hline
    Method & Gaussian Noise \\\hline
    REALM & \textbf{26.1}\\
    - $S_{\textrm{div}}$ & 9.2\\
    \hline
    \end{tabular}
    \caption{Accuracy for ResNet-50 GN evaluated on gaussian noise corruptions clarifying the importance of the $S_{\textrm{div}}$ term. Results are averaged over 3 runs.}
    \label{tab:sdiv}
    }
\end{table}

We find that the $S_{\textrm{div}}$ term is crucial for adaptation performance of REALM.  We hypothesize this is because we are no longer skipping samples based on reliability.  As such, we see many additional updates which have similar predictions.  This means $S_{\textrm{div}}$ will down-weight many of the samples we may not have used to update the model prediction EMA in other approaches.    

\begin{figure}[h!]
    \centering
    \includegraphics[width=2in]{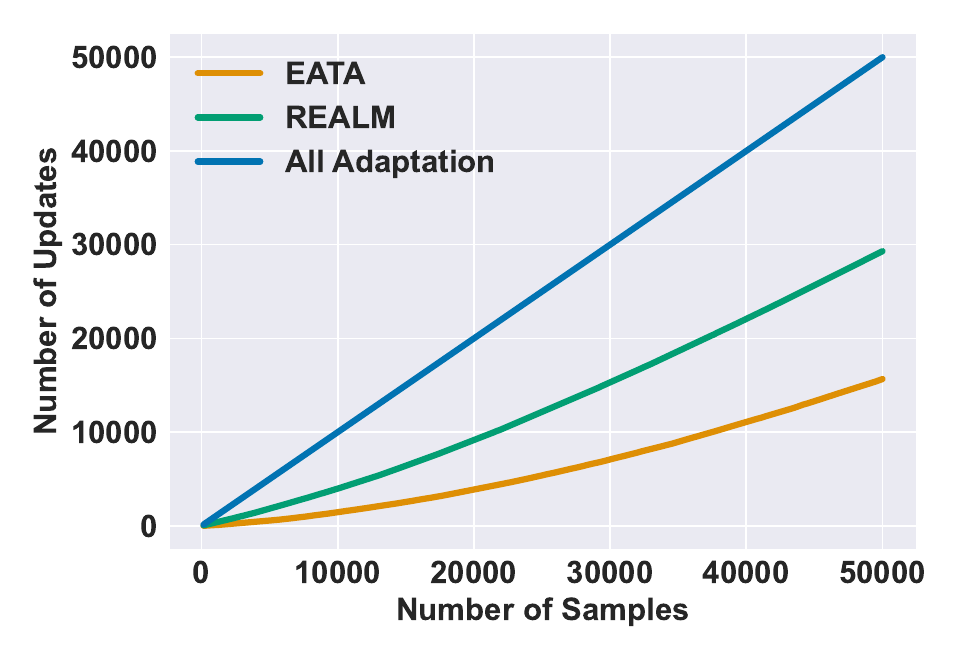}
    \caption{Number of updates for ResNet-50 GN evaluated on gaussian noise corruptions illustrating that REALM updates on twice as many samples as EATA.}
    \label{fig:realm_updates}
\end{figure}

We further show that even with the greater impact $S_{\textrm{div}}$ has on learning in REALM, we still find that REALM updates over more samples than EATA.  On ImageNet-C with gaussian noise corruption at severity 5, we find that REALM updates on twice as many samples as EATA in \cref{fig:realm_updates}.  We believe this leads to REALM achieving higher accuracy.

\subsection{Optimizing Scaled Squared Entropy}
Our choice of $\rho$ is modified from the general robust loss function defined in \cite{barron2019general}.  However, that loss function makes an implicit assumptions that the loss itself is a scaled squared norm: $\left(\frac{x}{\lambda}\right)^2$.  Traditionally robust loss functions are used in robust regression settings where the scaled squared norm is  suitable choice of loss.  However, in other settings such as here, minimizing the squared entropy is unconventional and an unmotivated objective.  Nonetheless, we show in \cref{tab:regression} that REALM still produces the same results on the gaussian noise corruption at severity 5 as this optimization has a similar form to that of implicit SPL.  

\begin{table}[ht]
    {\small
    \centering
    \begin{tabular}{|l|c|}
    \hline
    Method & Gaussian Noise \\\hline
    REALM & 26.1\\
    $\textrm{REALM}^2$ & 26.1\\
    \hline
    \end{tabular}
    \caption{Accuracy for ResNet-50 GN evaluated on gaussian noise corruptions comparing our modified robust loss function with the version used for regression problems. Results are averaged over 3 runs.}
    \label{tab:regression}
    }
\end{table}

\begin{figure*}
  \centering
  \begin{subfigure}{0.24\linewidth}
    \includegraphics[width=\columnwidth]{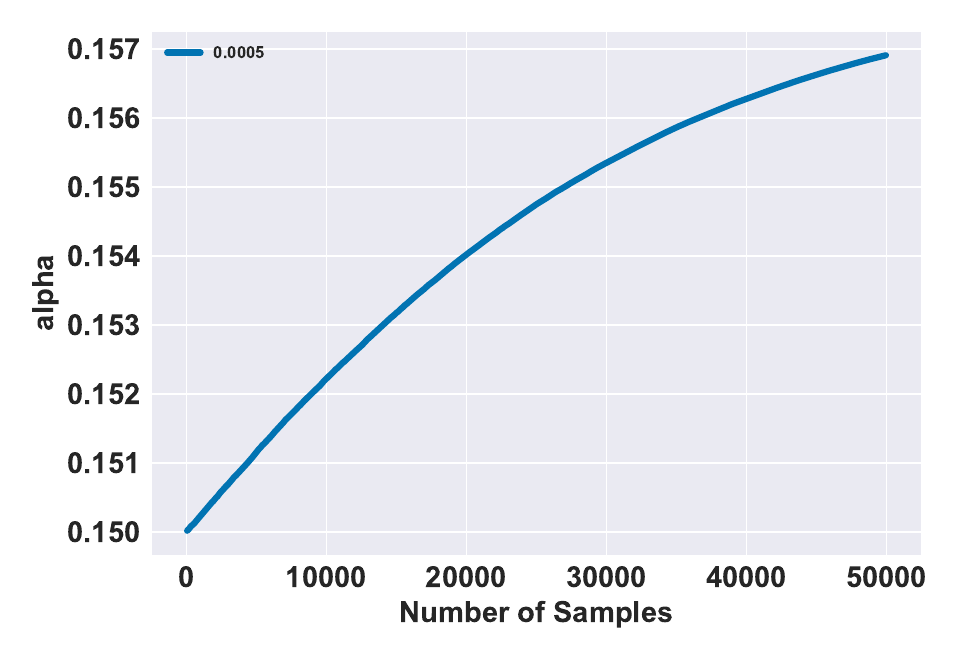}
    \caption{$\alpha$ - $\textrm{lr}=5e-4$}
  \end{subfigure}
  \hfill
  \begin{subfigure}{0.24\linewidth}
  \includegraphics[width=\columnwidth]{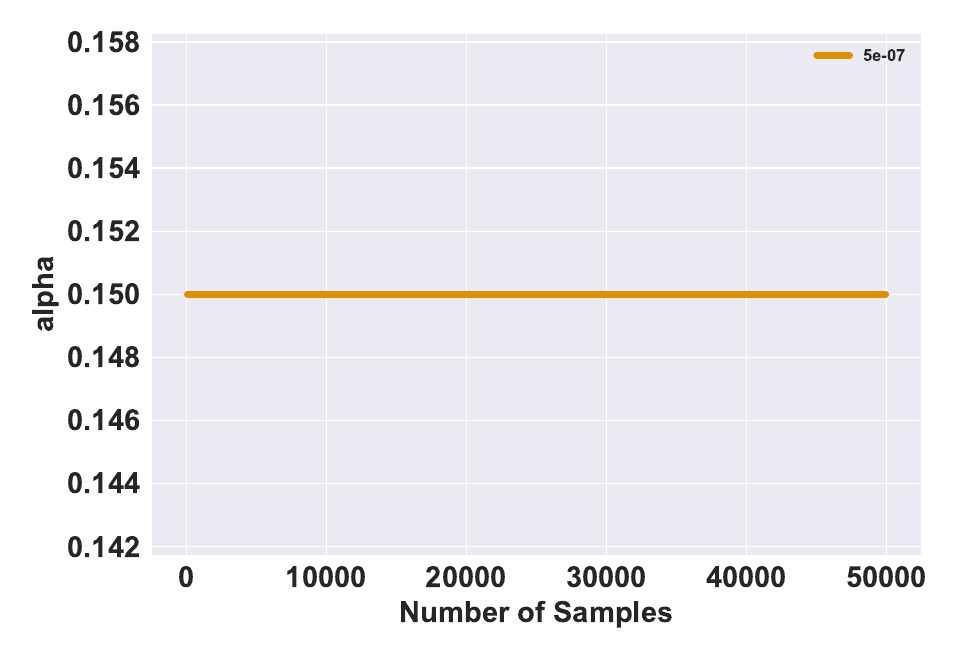}
    \caption{$\alpha$ - $\textrm{lr}=5e-7$}
  \end{subfigure}
  \begin{subfigure}{0.24\linewidth}
    \includegraphics[width=\columnwidth]{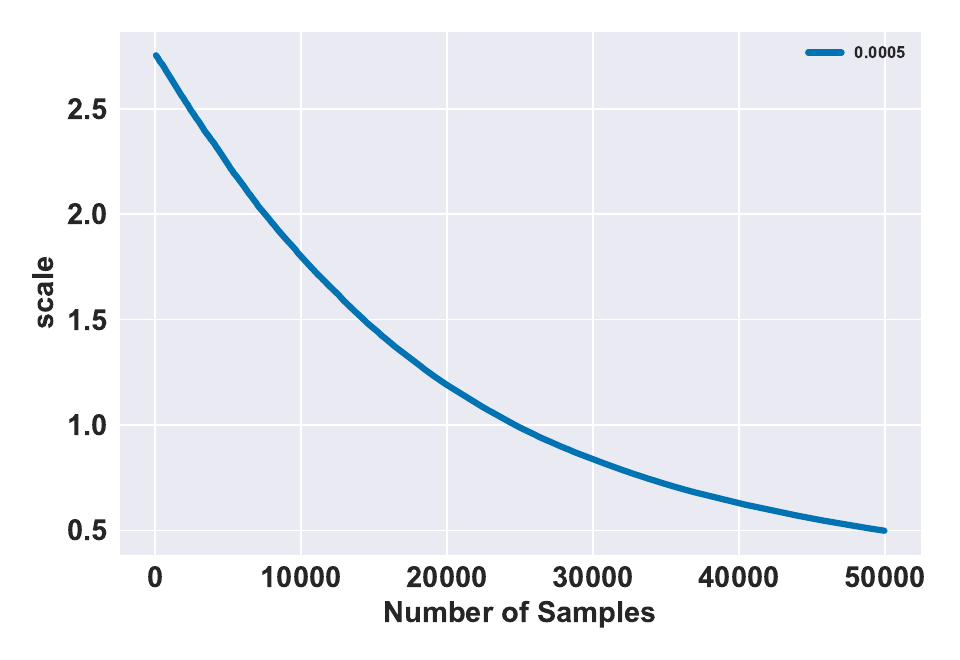}
    \caption{$\lambda$ - $\textrm{lr}=5e-4$}
  \end{subfigure}
  \hfill
  \begin{subfigure}{0.24\linewidth}
  \includegraphics[width=\columnwidth]{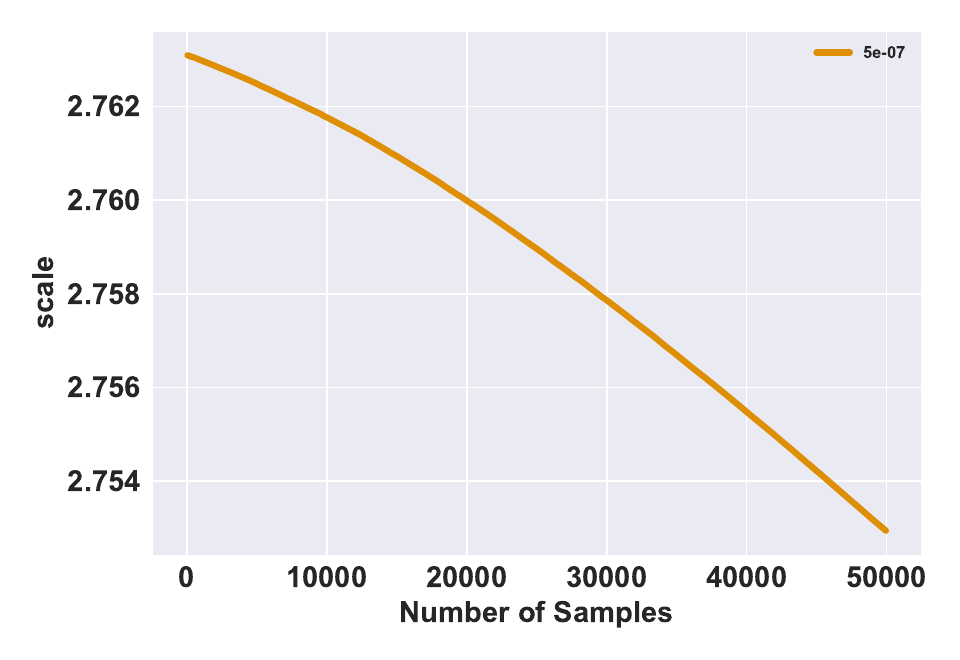}
    \caption{$\lambda$ - $\textrm{lr}=5e-7$}
  \end{subfigure}
  \caption{$\alpha$ and $\lambda$ during adaptation for the robust loss function.  Figures are over a single run to illustrate parameter changes.}
  \label{fig:params}
\end{figure*}

\subsection{Removing Loss Scaling}
The loss function reported in \cite{barron2019general} scales the gradient by a function of the loss scale $\lambda$.  For a given update, this lack of re-scaling entails that the the gradient will also be scaled by $\lambda$.  We modified the loss function to account for this scaled gradient and in \cref{tab:scale_factor} find that this term is crucial as it increases performance on the gaussian noise corruption by $4\%$.
\begin{table}[ht]
    {\small
    \centering
    \begin{tabular}{|l|c|}
    \hline
    Method & Gaussian Noise \\\hline
    REALM & \textbf{26.1}\\
    - Scale Factor & 22.0\\
    \hline
    \end{tabular}
    \caption{Accuracy for ResNet-50 GN evaluated on Gaussian noise corruptions comparing REALM with and without the scale factor for scaling the gradient update. Results are averaged over 3 runs.}
    \label{tab:scale_factor}
    }
\end{table}

\subsection{Frozen Top Block}
In prior work, it is shown that deep layers in the network are more sensitive and important to preserve for the original model as they retain semantic information, whereas the shallow layers of the model are important for covariate shifts.  Thus, in our main results, following SAR, we do not update the last block of the ResNet architecture with REALM ({\tt layer4}).  We, however test whether we need to freeze blocks of the network in REALM.  \cref{tab:freeze_block} shows it is unnecessary, and performance increases by $\sim 1\%$ on the gaussian noise corruption.  Still, we believe it may be important to limit model updates on higher severity corruptions as this can negatively impact adaptation, and potentially increase model forgetting on the in-distribution data.

\begin{table}[ht]
    {\small
    \centering
    \begin{tabular}{|l|c|}
    \hline
    Method & Gaussian Noise \\\hline
    REALM & 26.1\\
    - Frozen Block & \textbf{27.2}\\
    \hline
    \end{tabular}
    \caption{Accuracy for ResNet-50 GN evaluated on Gaussian noise corruptions comparing REALM with freezing the last block vs. all block norm layer updates. Results are averaged over 3 runs.}
    \label{tab:freeze_block}
    }
\end{table}

\subsection{Sensitivity to Learning Rate}
REALM is an extension of other entropy minimization methods that rely on sample skipping by reformulating as a SPL objective.  When recasting our objective, we have a few additional hyperparameters, most notably the learning rate on $\alpha$ and $\lambda$ which controls how much these parameters update. To keep comparisons between REALM and SAR consistent, we did not tune this value in the main text and set this value at $\textrm{lr}_{\alpha, \lambda} = 2 \times \textrm{lr}_{\theta}$.  Alternatively setting this directly to $ \textrm{lr}_{\theta}$ yields improved performance as well. In either case, we do not consider tuning  $\textrm{lr}_{\alpha, \lambda}$ in the main text as we did not want to give REALM any unfair advantage.  However, we find empirically in \cref{tab:small_lr} that  better performance using a smaller $\textrm{lr}_{\alpha, \lambda}$ can be achieved, however we were not able to tune for this value without a proper validation set.

\subsection{Visualizing $\alpha$ and $\lambda$}
Finally, we plot $\alpha$ and $\lambda$ for the robust loss during training in \cref{fig:params}.  We find that $\alpha$ increases or stays constant during training, while $\lambda$ decreases.  Both trends follow from the loss and adaptation behavior, since as we adapt during inference,  the loss on samples from the new distribution will gradually decrease on average, meaning $\alpha$ should increase for reduced penalization, and $\lambda$ should decrease for more gradient update.